\documentclass[pmlr,twocolumn,10pt]{jmlr} % W&CP article

\usepackage{booktabs}
\usepackage{siunitx}

\usepackage[switch]{lineno}

\usepackage{mathtools}
\usepackage{amsmath}
\usepackage{amssymb,amsmath}%,amsthm}
\usepackage{bbm}
\usepackage{xcolor}
\usepackage{hyperref}
\usepackage{gensymb}
\usepackage{algorithm}
\usepackage{algorithm2e} 
\usepackage{algpseudocode}
\usepackage{booktabs}
\usepackage{tikz}
\usetikzlibrary{arrows.meta, positioning, shapes.geometric}
\newcommand{\indep}{\perp \!\!\! \perp}
\usepackage[normalem]{ulem}

\usepackage{xcolor}
\definecolor{darkteal}{RGB}{0, 102, 102} % Dark teal (not too bright)

\definecolor{mutedpurple}{RGB}{102, 51, 153} % Muted purple

\theorembodyfont{\upshape}
\theoremheaderfont{\scshape}
\theorempostheader{:}
\theoremsep{\newline}

\jmlrvolume{LEAVE UNSET}
\jmlryear{2025}
\jmlrsubmitted{LEAVE UNSET}
\jmlrpublished{LEAVE UNSET}
\jmlrworkshop{Conference on Health, Inference, and Learning (CHIL) 2025} % W&CP title

\title[ExOSITO: Off-Policy learning for ICU lab test ordering]{
ExOSITO: Explainable Off-Policy Learning with Side Information for Intensive Care Unit Blood Test Orders}
\author{%
\Name{Zongliang Ji}\textsuperscript{1,3} \Email{jerryji@cs.toronto.edu} \\
\Name{Andre Amaral}\textsuperscript{1,2} \Email{AndreCarlos.Amaral@sunnybrook.ca} \\
\Name{Anna Goldenberg}\textsuperscript{1,3} \Email{anna.goldenberg@utoronto.ca} \\
\Name{Rahul G. Krishnan}\textsuperscript{1,3} \Email{rahulgk@cs.toronto.edu} \\
\addr \textsuperscript{1}University of Toronto, Canada \\
\addr \textsuperscript{2}Sunnybrook Health Sciences Centre, Canada\\
\addr \textsuperscript{3}Vector Institute, Canada \\
}

\begin{document}

\maketitle

\begin{abstract}
Ordering a minimal subset of lab tests for patients in the intensive care unit (ICU) can be challenging. Care teams must balance between ensuring the availability of the right information and reducing the clinical burden and costs associated with each lab test order. Most in-patient settings experience frequent over-ordering of lab tests, but are now aiming to reduce this burden on both hospital resources and the environment.
This paper develops a novel method that combines off-policy learning with privileged information to identify the optimal set of ICU lab tests to order. Our approach, EXplainable Off-policy learning with Side
Information for ICU blood Test Orders (ExOSITO) creates an interpretable assistive tool for clinicians to order lab tests by considering both the observed and predicted future status of each patient. 
We pose this problem as a causal bandit trained using offline data and a reward function derived from clinically-approved rules; we introduce a novel learning framework that integrates clinical knowledge with observational data to bridge the gap between the optimal and logging policies. 
The learned policy function provides interpretable clinical information and reduces costs without omitting any vital lab orders, outperforming both a physician's policy and prior approaches to this practical problem.
%This is the abstract for this article. If you are making your code
% available, \emph{do not link to it in the abstract since many indexing
% services will automatically remove or redact the link}. Instead,
% we are requiring every paper to have an initial statement on data and
% code availability right after the abstract.
\end{abstract}

\paragraph*{Data and Code Availability}
This paper uses the MIMIC-IV~\citep{johnson2023mimic} and HiRID~\citep{hyland2020early} datasets, both available on the PhysioNet repository. Code to reproduce the experimental results for ExOSITO is available at this repository\footnote{\url{https://github.com/Jerryji007/ExOSITO-CHIL2025}}.

\paragraph*{Institutional Review Board (IRB)} This study does not require IRB.

\section{Introduction}

\begin{figure*}[ht]
\begin{center}
\centerline{\includegraphics[width=0.85\textwidth, height=5.5cm]{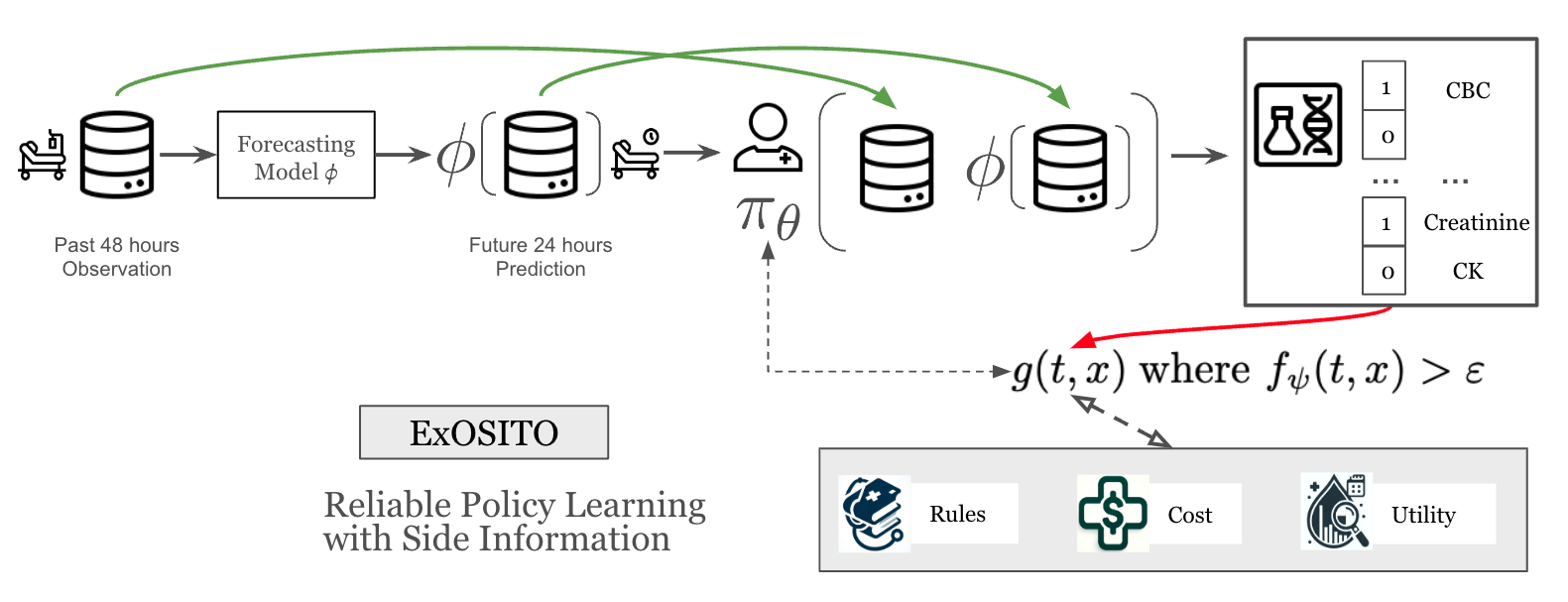}}
\caption{\small Overview of Proposed Method. Top: Development of an ICU patient status forecasting model $\phi$ for future predictions. Both observed past and predicted future of patient are inputs of policy $\pi_\theta$ (green arrow), which determines the next-day lab test order for the patient.
Bottom: $\pi_\theta$ is learned by maximizing the reward function $g$, which evaluates each test order (red arrow) and incorporates three main components (dashed arrow) along with $f_\psi$, the learned global propensity score (GPS) function, to ensure non-trivial support.
}
%\todo{explain red and green}
\label{fig:overview}
\end{center}
% \vskip -0.3in
\end{figure*} 

Ordering lab tests is a critical yet challenging task for clinicians, essential for assessing a patient's status and determining the appropriate treatment plan~\citep{test1997zimmer, kumwilaisak2008effect, ccmtest2023hjortso}. However, studies have shown that clinicians and hospitals often over-order lab tests \citep{feldman2009managing,badrick2013evidence}. This occurs for several reasons: a sense of responsibility, the mental ease of ruling out conditions, and pressures within medical hierarchies~\citep{griffith1997does, van1998we, korenstein2018impact,zhi2013landscape, sedrak2016residents}.
Over-ordering generates significant costs for patients, hospitals, and the environment while causing discomfort from unnecessary blood draws, and it may even worsen patient outcomes~\citep{berenholtz2004eliminating, salisbury2011diagnostic}. Moreover, redundant tests do not necessarily improve diagnostic accuracy~\citep{iosfina2013implementation, pageler2013embedding}. In this work, we propose a novel, explainable tool to assist clinicians in ordering the optimal set of lab tests for intensive care units (ICU) patients.

There are two categories of prior approaches; both face significant challenges that prevent real-world clinical deployment. The first category of work leverages \textbf{rule-based interventions}, such as capping the number of tests or reducing tests frequency~\citep{MayaDewan2016ReducingUP, dewan2017reducing, kotecha2017reducing}. %\textcolor{red}{
These strategies lack adaptability as they do not account for individual patient characteristics or changing clinical conditions over time.
This may lead to the omission of necessary tests and compromise care~\citep{kobewka2015influence}.
Another line of work ~\citep{cheng2018optimal, chang2019dynamic, ji2024measurement} uses offline, off-policy \textbf{reinforcement learning (RL)}. While these methods offer more adaptive decision-making by optimizing test ordering based on historical data, they rely on objectives that do not align with decision-making dynamics in clinical settings.
%While these methods offer more adaptive decision-making by optimizing test ordering based on historical data, they face limitations in handling the uncertainty and variability of real-time clinical settings.
First, they aim to replace clinical oversight -- a non-starter in practical deployments for high-risk medical settings such as ICUs since the legal and medical imperative for decision lies firmly with clinicians ~\citep{lu2020deep}. Our strategy instead is to focus on creating assistive tools for clinicians within the ICU. 
This requires reframing the learning problem from developing an RL agent meant to replace the clinician to a bandit formulation that offers real-time assistive support to the attending physician.
%This requires reframing the learning problem from obtaining an RL agent intended as a replacement for a clinician to a bandit formulation that provides instantaneous assistive help to the current attending physician.
Next, prior methods rely solely on historical data for decision making, making them susceptible to learn policies that are not robust to changing standards of care; e.g. some hospitals may have previously allowed more frequent test orders than current practices permit, leading to significant discrepancies between the offline policy and the intended real-world deployment scenario. 
This results in low-overlap between the offline and online policy. Finally, a significant concern during deployment is interpretability which is crucial for clinicians to build trust when using the tool during silent deployment. 
These challenges motivate our work. 
%There is consequently a need for new methodology to create more robust and practically useful methods to learn policies.
%Our aim is to develop an explainable and reliable method that integrates both data and clinical knowledge, serving as a``co-pilot" to help the clinical team decide the optimal lab tests to order for each patient.

%We introduce the first clinician-facing method for ordering blood tests for ICU patients.
%, providing pre-computed, safe recommendations for the optimal set of tests to order each day. Our approach aims to reduce cognitive load, lower costs, improve resource allocation, and ultimately enhance patient outcomes. 
%We demonstrate that our method outperforms both physician decision-making and previous approaches in lab test ordering across two real-world ICU datasets.
%In this work, we ask the question, if deployment is the goal, what is the learning tool that would provide the most clinical value to a doctor? Answering this question is marred with technical challenges. Different hospital environments use different policies for administering lab-tests; this motivates a need to create methodology that leverages global 

\textbf{Contributions:} Our strategy for improving the robustness of learned policies from offline data is to leverage \emph{physiologically grounded} privileged information (information that is available at training time, but not at test time) in the form of clinical rules. This enables us to mitigate the difference between the optimal and logging policy. To improve explainability, we provide practitioners a proxy of the learner's assessment of the \emph{current} patient state through the use of forecasts that visualize the expected future evolution for the patient.
%Our approach, EXplainable Off-policy learning with Side Information for ICU blood Test Orders (ExOSITO) thus creates an interpretable assistive tool for clinicians to order lab tests by considering both the observed and predicted future status of each patient. 

By combining these ideas we propose EXplainable Off-policy learning with Side Information for ICU blood Test Orders (ExOSITO); an interpretable assistive tool for clinicians to order lab tests using both the observed and predicted future status of a patient. To the best of our knowledge, this is the first work to use privileged information to create an explainable, clinician-facing tool for lab test ordering, leveraging verified clinical rules as side information within an off-policy bandit framework.
We study the framework on real-world medical datasets and showcase how the hybrid framework outperforms both a physician's policy and prior approaches bringing us significantly closer to a deployable clinical decision support tool.  
\section{Related Work}
% \vskip -0.2in
\emph{Machine Learning for Laboratory Test Ordering: }
While there have been some rule-based attempts to reduce redundancy in laboratory test ordering, particularly in pediatric and cardiac surgical ICU settings~\citep{MayaDewan2016ReducingUP, dewan2017reducing, kotecha2017reducing}, these methods do not generalize to other contexts. 
%Thus, there is a pressing need for data-driven approaches for the optimal scheduling of lab tests.
\citet{badrick2013evidence} employed a binary classifier to assess whether a given test contributes to information gain in the clinical management of patients with gastrointestinal bleeding. \citet{soleimani2017scalable} used regression methods to model the novel information each test provides. These approaches focus on specific diseases and  overlook a large number of clinical factors, such as predictive information from vital signs and the causal links between test orders and their utility to clinicians. Such omissions make the model less adaptable to variations in patient conditions and unsuitable for deployment in an ICU.

Off-policy learning, particularly via RL, has been extensively explored for ICU patient treatment plans~\citep{komorowski2018artificial, tang2022leveraging, ma2023learning, nambiar2023deep, emerson2023offline, kondrup2023towards, schweisthal2023reliable}. 
However, such methods typically aim to maximize patient outcomes as the reward (See Figure~\ref{fig:graphs}). The challenge with this formulation is that lab test orders should be viewed as actions that provide information to clinicians rather than as having a direct impact on patient outcomes. Ignoring the clinician as the key decision maker in the loop breaks the assumption underlying the Markov decision process. 
This motivates us to adopt a conceptually simpler bandit formulation aimed at providing instantaneous assistance to clinicians in deciding which lab tests to order.
%Prior work~\citep{cheng2018optimal, chang2019dynamic, ji2024measurement} has explored policy learning related to ICU patient measurements or lab tests. 
\citet{cheng2018optimal} focused on sepsis-related lab tests using only four lab features limiting its generalizability to broader ICU settings. \citet{chang2019dynamic} applied deep Q-learning to frequently measured features, such as vital signs but their work omits ordering for several blood test features. \citet{ji2024measurement} built on \citet{chang2019dynamic}'s framework by applying updated learning methods but found no significant differences across learned policies due to limitations in the logging policy and the use of mortality as the sole reward.

\emph{Off-Policy Learning with Constraints and Side Information: } Off-policy learning refers to learning a strategy for making decisions, called a policy, using data that was collected under a different task and/or goals.~\citep{lange2012batch, sutton2018reinforcement, levine2020offline}. %Given that our approach utilizes previously collected ICU patient EHR data to learn a policy for lab test ordering, it aligns well with off-policy learning.
Off-policy learning, without further interaction with the environment, is challenging, as the policy the data exhibits may not be optimal. This issue is evident in lab test ordering, where clinicians err on the side of ordering redundant tests. Previous studies have shown that learned policies can perform poorly with out-of-distribution actions \citep{fujimoto2019off}. To address this, some methods constrain the learned policy to actions within the support set of the behavior policy, either by introducing implicit regularization in the form of generative models to estimate the behavioral policy~\citep{zhou2021plas, ghasemipour2021emaq} or by incorporating regularization penalties to constrain the divergence between the learned and behavior policies~\citep{jaques2019way, kumar2019stabilizing, wu2019behavior, wu2022supported, mao2023supported}. 
Peripherally related is the field of Bayesian Optimization~\citep{yang2022offline, wilson2014using, letham2019bayesian, muller2021local, cheng2018optimal} which blends active and policy learning with the goal of maximizing an expected reward. However, Bayesian Optimization requires access to a simulator and precise simulators are  rarely available in healthcare. 
\citet{le2019batch} studies learning such policies with multiple constraints, and \citet{schweisthal2023reliable} leverages ideas from causal inference; specifically, propensity score estimation to guide the design of policies to ensure non-trivial overlap over covariate-treatment space. Our work builds on these ideas.  

%While extensive literature exists on ensuring the support of the behavior policy, the information is still extracted only from the observed dataset.
%The wealth of medical knowledge accumulated from decades of critical care medicine provides valuable guidance on policy learning. %Side information are rules, knowledge, or assumptions in addition to the required policy learning components.
%While some \emph{online} bandit algorithms leverage side information~\citep{slivkins2011contextual, kleinberg2019bandits, tang2023counterfactual}.
Side information has been used in off-policy evaluation~\citep{felicioni2022off} as well inverse reinforcement learning~\citep{wen2017learning} but not in a deployment focused application.
%There are works in the  In the off-policy evaluation and inverse RL  \citet{wen2017learning} and \citet{felicioni2022off} have utilized action-related side information. 
%To the best of our knowledge, there has not been a clinically actionable application of side information in the context of decision making in healthcare. 
In this work, we showcase the impact that even simple rules summarized by clinicians can, as side information, guide policy learning.
% \vskip -0.3in

\section{Background} \label{sec:problem}
% \vskip -0.2in
We model our problem as an off-policy \textbf{causal contextual bandits} problem with multi-dimensional binary actions. Given an observational dataset $\mathcal{D} = \{x_i, t_i, y_i\}_{i=1}^n$ with $n$ i.i.d ICU patient stays.\footnote{Here each ICU stay $i$ is considered a `time step' under standard bandits setting.}
$x_i \in X \subseteq \mathbb{R}^{d \times L}$ is the context, covariate, or a patient ICU stay represented by an irregular time-series matrix, where $d$ is the number of features and $L$ is the length of stay. 
$t_i \in T \subseteq \{0,1\}^K $ is the action, or the lab test order represented by a $K$-dimensional binary vector.
$y_i \in Y \subseteq \mathbb{R}$ is the reward, outcome, or the utility of the lab test order represented by a real number. We highlight that $T$ \emph{does not} refer to the drugs or clinical interventions that a patient may be prescribed within the ICU -- one of the assumptions that we make in this work is that the clinical covariates, $X$, capture the effect of any medication/clinical intervention and consequently sufficient evidence to decide upon which lab test to order. 

Clinicians are keen to find a policy $\pi : X \rightarrow T$, which determines the lab tests to order for the following day given patient status as context (covariates). 
The policy value, $V(\pi) = \mathbb{E}_\pi [Y(\pi(X))]$, is the expected reward (outcome) of policy $\pi$. 
The function,  $Y(\cdot):T \rightarrow \mathbb{R}$, measures the potential outcome given actions (treatments).
Our objective is to find optimal policy $\pi^*$ from policy class $\Pi$ that maximizes the policy value $V(\pi)$, as expressed in the following equation:
% \vskip -0.2in
\begin{equation} \label{eq:prob_baseobjective}
\pi^* \in \underset{\pi \in \Pi}{\arg\max}\; V(\pi).
\end{equation}
% \vskip -0.15in
As it is understood that clinicians' planning and actions vary, we assume that the logging policy represents a sub-optimal policy within the set $\Pi$.
% Since our problem is under the off-policy setting, no timely feedback on how much lab test orders affect clinicians, $Y(\cdot)$ is not available. 
Since our problem operates under an off-policy setting, timely feedback on $Y(\cdot)$, which represents the effects of lab test orders on clinicians, is not available.
In our method, we use the \textit{conditional potential outcome function}, $g(t, x) = \mathbb{E} [Y(t) \,\mid\, X=x]$, to estimate individual potential outcome as a proxy of $Y(t)$. This makes our policy value, $V(\pi) = \frac{1}{n} \sum_{i=1}^n g(t_i, x_i)$, when evaluating the policy $\pi$ on $n$ samples. Prior work models $g$ as patient mortality estimation, which, while easily obtainable, is not informative for our problem~\citep{chang2019dynamic, schweisthal2023reliable}. Our work creates a novel variant of $g(t,x)$, a closed-formed differentiable function that we refer to as the \textit{lab order utility function}. 

\textbf{Assumptions:}  
%\teal{We assume that the drugs prescribed to patients (over-time) within the ICU that are not included in our covariates do not influence the choice of which lab test to order. } 
% i.e. their influence is mediated entirely by the observed clinical covariates at that point in time. 
We assume that the missingness pattern that the longitudinal data exhibit is missing at random. For the outcome estimation to be identifiable, we adhere to three standard assumptions in causal inference \citep{rubin1974estimating}: 
(1) \emph{Consistency} ($Y = Y(T)$ \footnote{Abusing notation here, function $Y(\cdot)$ means the potential outcome and $Y$ means actual outcome.}), asserting that observed outcomes align with potential outcomes under the observed treatment. 
(2) \emph{Ignorability} ($Y(t) \; \indep \; T \,\mid\, X, \;\forall t \in T$), confirming the absence of hidden confounders \footnote{We assume that our set of covariates, which includes treatments (for example drugs and procedures), vital signals, and test results, is complete. A detailed discussion on the validity of this assumption can be found in Appendix~\ref{apx:ignore_assumption}.}. 
(3) \emph{Overlap} ($f(t, x) > \varepsilon$, $\forall x \in X, t \in T$, for some $\varepsilon \in \left[0, \infty \right)$), ensuring all potential treatments can be accurately estimated for every individual. Here, $f(t,x) = f_{T\,\mid\, X=x}(t)$ is the \emph{global propensity score} (GPS) represents the conditional density of $T$ given $X=x$. %\footnote{GPS measures the probability of certain lab test $t$ being ordered given patient ICU stay $x$.}.
We differentiate between \emph{weak} overlap ($\varepsilon = 0$) and \emph{strong} overlap ($\varepsilon > 0$), focusing predominantly on \emph{strong} overlap due to its enhanced reliability in finite sample contexts, unless specified otherwise.

\textbf{Limited Overlap}: Restricted overlap introduces both empirical and theoretical hurdles. Firstly, datasets with high dimensionality or limited sample sizes often experience sparse coverage in the $X \times T$ space, leading to reduced overlap \citep{d2021overlap}. This limitation increases uncertainty in our lab order utility function $g(t,x)$, hindering effective decision-making. Secondly, the possibility of small $\varepsilon$ values in certain areas (due to unobserved patient trajectories or specific unassigned tests) necessitates a dependable off-policy approach.
% \vskip -0.1in
%\textbf{Addressing Limited Overlap}: Followed from~\citet{schweisthal2023reliable}, we confront 
Our work tackles this challenge by leveraging GPS. However, this is often not directly available and must be inferred from observational data. Our approach involves estimating the GPS as a probability density function $f(t, x) = f_{T\,\mid\,X=x}(t)$, correlating lab test orders $T$ with the patient's ICU stay $X=x$. Following ~\citet{schweisthal2023reliable}, we opt for \emph{conditional normalizing flows} (CNFs) \citep{trippe2018conditional,winkler2019learning} to form a parametric estimate of the GPS function. CNFs, built on the foundation of normalizing flows \citep{tabak2010density, rezende2015variational}, are parametric generative models capable of modeling conditional densities $p(y|x)$. CNFs are able to transform a simple base density $p(z)$ through an invertible transformation, parameterized by $\gamma(x)$, dependent on the input $x$ which makes CNFs suitable for density estimation.
The training of CNFs is guided by minimizing the negative log-likelihood loss, $ \mathcal{L}_{\text{nll}}= - \frac{1}{n}\sum_{i=1}^n  \log  \hat{f}(t,x)$\footnote{Additional details on the CNF training process are provided in the Appendix~\ref{apx:cnf}.}. Our method for policy learning uses the learned GPS function $\hat{f}(t,x)$,  steered away from areas of high uncertainty, improving the reliability of the resulting policy.

\section{Explainable Off-policy learning with Side Information for Test Orders (ExOSITO)} 
% %\vskip -0.2in
%\todo{Name for the method.... hard}
% \begin{itemize}
%     \item we find policy with explainability, later on can be accessible by experts for ‘online’ approval, OR adding the predicted future can provide clinician a chance to better evaluate learned policy suggestions
%     \item Clinician planned (rules) and acted (observation) are different, and our dataset is not perfect, we mitigate this by this outcome/reward function that has multiple terms
%     \item there is a overlap issue in the collected dataset, we mitigate that by applying other’s (the neurlps paper) method using propensity score as a training constraint
% \end{itemize}

%\todo{add another paragraph in (done)}
% Clinicians planned and acted differently, 
% they are trained on one thing 
% but when they have loaded amount of patients they would go with the easiest 'order everything'
% they also feel more responsible when ordering more
% however, the basic rules and knowledge summarized from years of practice could be a second opinion for them to allow them to have a second thought on the ordering

Clinicians are trained through decades of formal education and experience, planning their actions based on established practices. However, in real clinical settings, especially under the pressure of high patient loads, clinicians may resort to ordering as many tests as a precautionary measure, often driven by a heightened sense of responsibility. By incorporating basic rules and knowledge accumulated from years of practice, our approach offers a second opinion, encouraging clinicians to reconsider their initial test orders. This integration of foundational rules with clinical practice helps streamline decision-making and ensures more targeted and efficient patient care.
%%\vskip -0.1in

\textbf{Summary} We introduce an explainable approach to creating a learned policy for ordering laboratory tests in the ICU (refer to Figure~\ref{fig:overview}). 
If every clinician had the bandwidth to stop and think about what lab tests to order, then one of the decision thresholds they might use is whether or not they think the patient is on a trajectory to worsening or recovery. 
To leverage this insight we build a forecasting model to predict a patient's future physiological status using their laboratory biomarkers as a proxy for the same. 
In conjunction with a patient's immediate history, these predictions then form the inputs for our policy function. 
%, as they mimic clinicians' decisions on which labs to order based on both the observed patient condition and their intuition about the patient's future trajectory.
%These predictions are then utilized as inputs for our policy formulation. These predictions are necessary as clinicians decide which labs to order based on both observed patient condition and clinicians intuition on patients future trajectory.
Next, we establish bounds for each observed lab test order based on clinically validated guidelines. 
We identify the minimum sets of orders to make by applying clinician-curated rules, and determined the maximum sets of orders by combining the observed and rule-generated orders.
These rules are used in the design of a potential outcome function to evaluate the effectiveness of each potential lab test order. They help us mitigate the disparity between physician policy (i.e. logging policy) and our learned policy $\pi$. By using these rules as side information in the reward function, we can regularize the off-policy learning algorithm to ensure that it learns a safe and reliable policy for lab test ordering in ICU settings.
% %\vskip -0.8in

% \vskip -1.4in
%Context Setup}
%Multivariate Patient Status Forecasting
\subsection{Forecasting for explainability}
\vskip -0.1in
Previous studies~\citep{cheng2018optimal, chang2019dynamic, ji2024measurement} represent patient covariates $X$ as an imputed, irregular time-series matrix of the patient's history. Our interactions with medical experts revealed that when clinicians order lab tests with forethought, they attempt to infer insights into a patient's future condition; i.e. they focus not only on current patient states but also on forecasting future lab results when making their prognoses about the patient. This insight prompted the inclusion of both observed and predicted patient states in our representation of patient covariates. To predict patient future status, we first build our forecasting model, $\phi$, which is based on PatchTST~\citep{nie2022time}. $\phi$ takes the observed patient status $X_\text{prev}$ as input. This input comprises a broad spectrum of features, including vital signs, treatments, and relevant lab test results based on recommendations by ICU clinicians. The model is trained to minimize the mean squared error, $\mathcal{L}_\textit{mse} = \frac{1}{n} \sum_{i=1}^n (\hat{x}_{post} - x^*_{post})^2$, between the observed future status $X^*_\text{post}$ and the predicted future status $\hat{X}_\text{post} = \phi(X_\text{prev})$\footnote{Details about the construction of our patient status forecasting model are provided in Appendix \ref{apx:forecast}.}. By constructing the context $X$ with patient past and predicted future status, we can learn policies that are explainable to clinicians as each policy output correspond with exact values of patient status. This setup can provide clinicians a better chance to evaluate policy actions during future deployments for `online' approvals. 
\vskip -0.15in
%later on be accessible by medical experts for `online' approvals and 

\begin{algorithm}[tb]
\caption{Obtain bounds for observed test orders}
\label{alg:orderbound}
\small
\KwIn{Patient stay $x$, set of clinical rules $\mathcal{CR} = \{r^1,\ldots,r^M\}$}
\KwOut{Minimal $t^{min}$ and maximal $t^{max}$ of lab test order of stay $x$}
Determine observed test order $t^* \in \{0,1\}^K$ from $x$\;
$t^{max}, t^{min} \gets \vec{0} \in \{0,1\}^K$, where $K$ is the number of lab tests\;
\ForEach{$r^m \in \mathcal{CR}$}{
    \If{$x\ \text{satisfies}\ r^m$}{
         Find the indices $\mathcal{I} \subset \{1,\ldots,K\}$ of lab tests corresponding to rule $r^m$\;
         $t^{min}_i \gets 1$ for $i \in \mathcal{I}$\;
    }
    $t^{max} \gets \{t^*_j \lor t_j^{min}\}_{j=1}^K$\;
}
\end{algorithm}

%%\vskip -0.3in
\subsection{Minimal and maximal expectations of lab tests}
\vskip -0.1in
\label{boundgen}

% \vskip -0.5in

% \begin{algorithm}
% \caption{Obtain bounds for observed test orders}
% \label{alg:orderbound}
% \begin{algorithmic} %[1] % The [1] option enables line numbering
% \State \textbf{Input:} Patient stay $x$, Set of clinical rules $\mathcal{CR} = \{r^1, ..., r^M\}$
% \State \textbf{Output:} Upper bound $t^{upper}$ and lower bound $t^{lower}$ of lab test order of stay $x$
% \State Determine observed test order $t^* \in \{0,1\}^K$ from $x$
% \State $t^{upper}, t^{lower} \gets \vec{0} \in \{0,1\}^K$ where $K$ is number of lab tests \\
% \For{$r^m$ in $\mathcal{CR}$}
% \IF{$x\ \text{satisfies}\ r^m$}
% \State Find the indices $\mathcal{I} \subset \{1,...,K\}$ of lab tests corresponding to rule $r^m$
% \State $t^{lower}_i \gets 1$ for $i \in \mathcal{I}$
% \ENDIF
% \State $t^{upper} \gets \{t^*_j \lor t_j^{lower}\}_{j=1}^K$
% \EndFor
% \end{algorithmic}
% \end{algorithm}

Lab test ordering in ICU is a well-established practice, underpinned by decades of clinical experience, which has yielded straightforward guidelines for test ordering~\citep{kumwilaisak2008effect, cismondi2013reducing, vidyarthi2015changing, bindraban2018reducing}. For example, a clinician would typically order a Complete Blood Count (CBC) for a patient who has undergone a blood transfusion within the last 48 hours. Despite this, the choices made regarding which labs to order in the physician policy may not represent the optimal set. But why learn a policy when such guidelines exist? The answer is twofold: (1) These guidelines are basic, derived from historical knowledge, and tend to be exceedingly conservative, being triggered only when patients meet certain extreme criteria. (2) These guidelines apply to both past and future patient states. For example, if a clinician knows with certainty that a patient's hemoglobin will drop below a threshold tomorrow, a CBC should be ordered preemptively. However, clinicians cannot accurately predict future patient conditions in practice. Despite this, such guidelines are uniquely suited to be incorporated as priors in policy learning systems, as they are inherently conservative and can be combined with forecasting tools.

To effectively incorporate clinical guidelines into our policy learning framework, we derive stay-specific bounds that constrain the possible set of lab tests to be ordered.
%\teal{ 
For each patient stay $x$, we can determine a conservative bound $t^{min} \in \{0,1\}^K$ for an order of $K$ possible lab tests, based on these clinically derived guidelines applied to $x$. 
Considering our assumption that the observed treatment policy is suboptimal and possibly excessive~\citep{wang2017optimal, sachdeva2020off}, we establish a permissive bound of each test order as $t^{max} = \{t^*_j \lor t_j^{min}\}_{j=1}^K  \in \{0,1\}^K$, which is a combination of the guideline-derived order $t^{min}$ and the observed order $t^*$. 
This is because approximately 8\%-12\% of tests are missed in observed orders compared to guideline-based orders due to timing discrepancies or end-of-stay variations. 
The methodology for deriving these bounds is outlined in Algorithm~\ref{alg:orderbound}. 
For a rule stating, `If Hemoglobin is less than 7, order CBC.' For a given patient stay $x$, if a Hemoglobin measurement is less than 7, then $t^{min}_{CBC} = 1$.
Due the conservative nature of the clinical guidelines, the guideline-generated orders constitute about 30\% of the observed orders. Further details on the rule generation process and supporting literature for these clinical guidelines can be found in Appendix~\ref{apx:boundgen}. %}

%\vskip -0.6in

\subsection{Potential Outcome Function for Lab Test Order Utility} \label{deltax}
%\vskip -0.15in
Since clinicians plans (rules) and actions (observations) differ, the treatments (labs) in the collected dataset $\mathcal{D}$ is not perfect, we mitigate this disparity by defining a expected outcome function $g(t,x)$ with multiple terms.
%With our patient status forecasting model and the established lab test order bounds, we proceed to define the expected outcome function $g(t,x)$. 
This function assesses the utility of a lab test order $t$ in the context of a patient's status $x$. It also serves as a critical estimator for the policy value $V(\pi)$. 
Unlike previous studies~\citep{chang2019dynamic,schweisthal2023reliable} that primarily used mortality as a metric, our focus is on the usefulness of lab tests to clinicians rather than direct patient outcomes. This is due to the fact that lab tests principally aid clinicians in decision-making. Quantifying the exact usefulness of each lab test to clinicians is complex, but from our discussions with medical professionals, we identified three characteristics of an effective lab test order:

\emph{a] Informative:} The lab tests should provide maximum information to clinicians. That is, if a test $t_i$ is predicted with less variability than another test $t_j$, the necessity to order $t_j$ becomes more significant as high variability means patient status changed drastically. For a test order $t$ and patient status $x = [x_{\text{prev}}, \hat{x}_{\text{post}}]$ (comprising both observed past and predicted future statuses), the test result variation is defined as:
%\vskip -0.25in
\begin{equation}
\Delta X(t,x) = \Delta_{avg}(t,x) + \Delta_{range}(t,x).
\end{equation}
%\todo{explain why not Variance}
We calculate $\Delta_{avg}$ to gauge the mean variation of test values:
%\begin{equation}
$$\Delta_{avg}(t,x) = \sum_{j=1}^K \mathbbm{1}(t_j > 0.5) \cdot |\overline{x_{prev}^{lab, t_j} } - \overline{x_{post}^{lab, t_j} }|,$$
%\end{equation}
where $\overline{x^{lab, t_j}}$ signifies the average feature value corresponding to test $t_j$. The indicator term is used as $t \in [0,1]^K$ represents the predicted probability of each test being ordered.
$\Delta_{range}$ measures the variation in the extremities of the test values ordered:
%\begin{equation}
$\Delta_{range}(t,x) = \sum_{j=1}^K \mathbbm{1}(t_j > 0.5) \cdot \max(\delta_{max}, \delta_{min})$,
%\end{equation}
where $\delta_{max}=|\max(x_{prev}^{lab, t_j}) - \max(x_{post}^{lab, t_j})|$ and $\delta_{min} = |\min(x_{prev}^{lab, t_j})- \min(x_{post}^{lab, t_j})|$ are the absolute differences between the maximum and minimum values of predicted and observed values for test $t_j$. 
Our decision to focus on average and extreme value differences rather than variance followed clinical consultations, favoring metrics that assess whether features meet or exceed clinically meaningful thresholds, aligning closely with practical needs in real-world settings. The clinical and literature support of our design on $\Delta X$ is further illustrated in Appendix \ref{apx:deltax}.

\emph{b] Safe, yet useful:} Lab test orders are encouraged to be safe in terms of meeting conservative requires laid out by the rules, while should conform to defined in Sec~\ref{boundgen}. The deviation of each test order from the bounds $t_{j}^{{min}}$ and $t_j^{{max}}$ is quantified by $\mathcal{L}_b$:
% %\vskip -0.4in
\begin{equation} \label{eq:smoothbound}
    \mathcal{L}_b(t,x) =  \sum_{j=1}^K \mathbbm{1} (t_j^{{max}} = t_{j}^{{min}}) \cdot |t_{j}^{{min}} - t_j|.
\end{equation}
% %\vskip -0.25in
The indicator function ensures the policy includes all necessary labs suggested by rules while avoiding redundant ones.
A lower value of $\mathcal{L}_b$ indicates that the test order $t$ is better aligned with the criteria of meeting the minimal and maximal expectations for orders. Here we treat $t \in [0,1]^K$ as the predicted probability for each test for our policy.

\emph{c] Cost effective:} The objective includes minimizing the cost of the ordered tests, aiming to reduce redundancy and, consequently, the financial and environmental burden. Differing from previous studies~\citep{cheng2018optimal, chang2019dynamic, ji2024measurement} that assume uniform cost across tests, we consider the relative clinical costs of each test. The cost function is defined as:
% %\vskip -0.25in
\begin{equation}
C(t) = \sum_{j=1}^K \alpha_j \cdot \mathbbm{1}(t_j > 0.5), \sum \alpha_j = 1,
\end{equation}
% %\vskip -0.16in
where $\alpha_j$ represents the cost associated with each lab test.
%\todo{we found these price based on medical journals}
Synthesizing these desirable qualities, we define the lab test order utility function (conditional outcome function) as:
% %\vskip -0.2in
\begin{equation} \label{eq:gtx}
g(t,x) = \Delta X(t,x) - \beta_1 \mathcal{L}_b(t,x) - \beta_2 C(t),
\end{equation}
% %\vskip -0.13in
with $\beta_1$ and $\beta_2$ as regularization hyperparameters. Next, we show how we use the GPS function, $g(t,x)$ to estimate the policy value $\hat{V}(\pi)$ during policy learning. 
%From patient facing (mortality) to clinician facing (usefulness)
% %\vskip -0.15in

\begin{algorithm}[tb]
\caption{Reliable off-policy learning for ICU blood test ordering}
\label{alg:bounds}
\scriptsize
\KwIn{Data $(X, T, Y)$, reliability threshold $\overline{\varepsilon}$}
\KwOut{Optimal reliable policy $\hat{\pi}_{\theta}^\text{rel}$}
% \vspace{0.5em}

\tcp{Step 1: Learn a multi-variate time-series forecasting model to predict future stay}
Estimate $\phi(x)$ that predicts patient's next 24 hours based on the past 48 hours status\;
% \vspace{0.5em}

\tcp{Step 2: Find lab order bounds using Algorithm \ref{alg:orderbound} and define lab test utility function}
Prepare $t^{max}$, $t^{min}$ 
and {\tiny $g(t,x) = \Delta X(t,x) - \beta_1 \mathcal{L}_b(t,x) - \beta_2 C(t)$}
% \vspace{0.5em}

\tcp{Step 3: Estimate GPS using conditional normalizing flows}
Estimate $\hat{f}(t, x)$ via loss $\mathcal{L}_{\text{nll}}$\;
% \vspace{0.5em}

\tcp{Step 4: Train policy network using reliable learning algorithm}
$\pi_{\theta}^{(k)} \gets \text{initialize randomly}$\;

$\lambda \gets \text{initialize randomly}$\;
% \vspace{0.5em}

\For{$m=1$ \KwTo $M$}{
  \For{each epoch}{
    \For{each batch}{
      \tcp{Predict next 24 hours ICU stay with forecasting model}
      $x_{post} \gets \phi(x_{prev})$\;
      
      $x_i \gets [x_{prev}, x_{post}]$\;
      
      $\mathcal{L}_{\pi} \gets  -  \frac{1}{n}\sum_{i=1}^{n} \cdot \\ \Biggl\{ g \Bigl( \pi_{\theta}^{(m)}(x_i), x_i \Bigr) -  \lambda_{i} \Bigl[ \hat{f} \Bigl( \pi_{\theta}^{(m)}(x_i), x_i \Bigr) - \overline{\varepsilon} \Bigr] \Biggr\}$\;
      
      $\theta \gets \theta - \eta_{\theta}\, \nabla_{\theta} \mathcal{L}_{\pi}$\;
      
      $\lambda \gets \lambda + \eta_{\lambda}\, \nabla_{\lambda} \mathcal{L}_{\pi}$
    }
  }
}
% \vspace{0.5em}
\tcp{Select best learned policy with respect to constrained objective on validation set}
$\pi_\theta^\text{rel} \gets \pi_{\theta}^{(m^*)}$, with 
\[
m^*= \underset{m}{\arg\max}\sum_{i=1}^{n} g \Bigl( \pi_{\theta}^{(m)}(x_i), x_i \Bigr) \cdot \mathbbm{1}\Bigl\{ \hat{f} \Bigl( \pi_{\theta}^{(m)}(x_i), x_i \Bigr) > \overline{\varepsilon} \Bigr\}
\]
\end{algorithm}

\begin{figure*}[ht]
\begin{center}
\centerline{\includegraphics[width=0.75\textwidth, height=8cm]{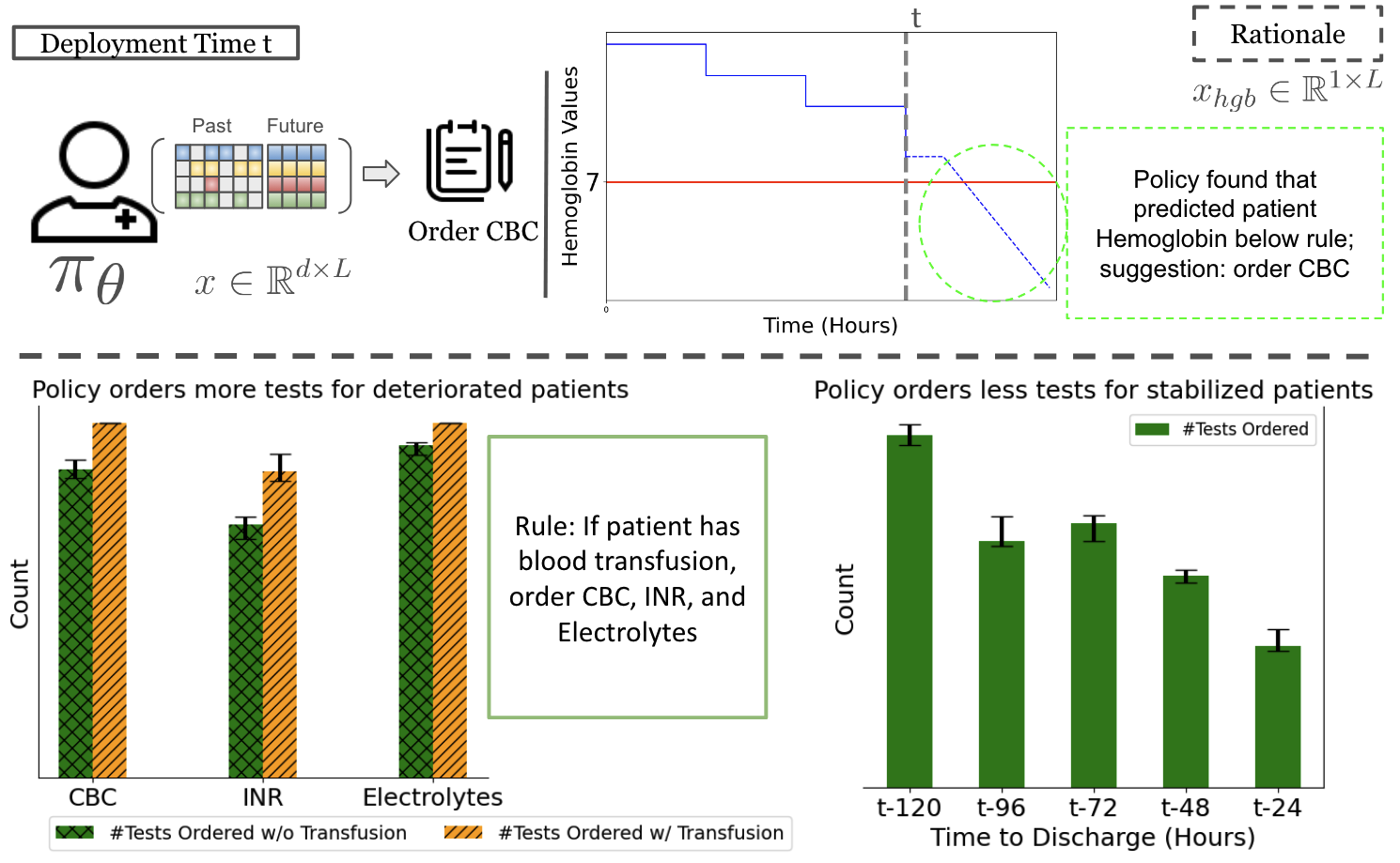}}
\caption{\small Integration of Medical Knowledge into Explainable Policy Learning for real patient data. Top: At deployment, our policy transparently justifies actions, such as ordering a CBC test, based on predictions like a decrease in future Hemoglobin levels. Bottom: The policy incorporates clinical guidelines for lab test ordering. 
At test time, the policy increases test orders for deteriorating patients due to changes in transfusion features, resulting in more corresponding tests being ordered (left). Conversely, it reduces test orders for stabilized patients as discharge approaches (right).}
\label{fig:rules}
\end{center}
%\vskip -0.45in
\end{figure*}

\subsection{Overlap-Guaranteed Policy Learning}
%\vskip -0.15in
We present a time-aware, overlap-guaranteed off-policy learning algorithm designed for ICU lab test ordering, prioritizing explainability and reliability. This algorithm
%is modified from \citet{schweisthal2023reliable}'s  
seeks to optimize policy $\pi^\text{rel}$ to maximize the estimated policy value $\hat{V}(\pi)$ while ensuring substantial data support in the covariate-treatment domain to guarantee the overlap condition.

The policy class $\Pi^{\mathrm{r}}= \left\{ \pi \in \Pi\ \,\mid\, f \left(\pi(x), x) > \overline{\varepsilon} \right), \; \forall x \in X \right\}$ is defined to include only those policies that maintain a minimum overlap, specified by a reliability threshold $\overline{\varepsilon}$. We then reformulate our objective from Eq.~\eqref{eq:prob_baseobjective} as:
%%\vskip -0.12in
$\pi^\text{rel} \in \underset{\pi \in \Pi^{\mathrm{r}}}{\arg\max} \; \hat{V}(\pi)$.
Given the finite nature of our observational data, we formulate the optimization as:
%\vskip -0.4in
\begin{equation}
\label{eq:method_constrainedobjective}
    \begin{aligned}
         & \underset{\pi}{\max} & \frac{1}{n}\sum_{i=1}^{n} g \left( \pi(x_i), x_i \right) & &
         \text{s.t.}  & & \hat{f} \left( \pi(x_i), x_i \right) \geq \overline{\varepsilon} &  \\
    \end{aligned}
\end{equation}
%\vskip -0.16in

Here the lab test order utility function $g$ serves as our policy outcome estimator, and the estimated GPS $\hat{f}$ limits the policy search space. To facilitate learning, we employ neural networks parameterized by $\theta$ and reformulate the constrained problem in Eq.~\ref{eq:method_constrainedobjective} into an unconstrained Lagrangian form:
%\vskip -0.4in
{\small
\begin{equation}\label{eq:lagrangian_optimization}
    \begin{aligned}
        \underset{\theta}{\min} \; \underset{\lambda_i \geq 0}{\max}  - \frac{1}{n}\sum_{i=1}^{n} \left\{ g \left( \pi_{\theta}(x_i), x_i \right) - \lambda_i \left[ \hat{f} \left( \pi_{\theta}(x_i), x_i \right) - \overline{\varepsilon} \right] \right\},
    \end{aligned}
\end{equation}}
where $\lambda_i$ are Lagrange multipliers.
%\vskip -0.1in

This adversarial learning approach, utilizing gradient descent-ascent techniques~\citep{Lin.2020}, enables the establishment of a robust policy under the defined constraints. The implementation specifics and algorithmic details are further elaborated in Algorithm~\ref{alg:bounds}.% \footnote{In \citet{schweisthal2023reliable}, $g(t,x)$ is a learned mortality predictor, whereas in our work, $g(t,x)$ is explicitly defined as a reward function. Furthermore, our algorithm includes an additional step that incorporates side information, diverging from prior approaches that rely solely on data from the logging policy.} in the Appendix~\ref{apx:pihyper}.

\begin{table*}[t]
\caption{Test set performance for baseline and our learned policies.}
% %\vskip -0.15in
\label{baselinepolicies}
\begin{center}
\begin{small}
\begin{sc}
\resizebox{\textwidth}{!}{
\begin{tabular}{l|ccccc|ccccc}
\toprule
  & \multicolumn{5}{c|}{MIMIC-IV} & \multicolumn{5}{c}{HIRID} \\
Policy & $\Delta X \uparrow$ & Cost$\downarrow$ & $\mathcal{L}_b^{test} \downarrow$ & $L_{low}\downarrow$ & $L_{up}\downarrow$ & $\Delta X \uparrow$ & Cost$\downarrow$ & $\mathcal{L}_b^{test} \downarrow$ & $L_{low}\downarrow$ & $L_{up}\downarrow$ \\
\hline
$\text{Random}_{0.5}$    & 0.23 & \textbf{0.51} & 4.3    & 3.2 & 1.1  & 0.47  & \textbf{0.49} & 4.4    & 2.9& 1.5 \\
$\text{Random}_{0.75}$   & 0.34 & 0.75 & 3.64   & 1.6 & 2.04 & 0.58  & 0.74 & 3.77   & 1.66 & 2.11 \\
LowerBound               & 0.37 & \textbf{0.62} & 0      & 0 & 0 & 0.62 & \textbf{0.35} & 0      & 0 & 0 \\
UpperBound               &\textbf{0.44} & 0.82 & 0      & 0 & 0 & \textbf{1.13} & 0.59 & 0      & 0 & 0 \\
Physician                & 0.41 & 0.67 & 1.24   & 1.24 & 0 & 0.96& 0.55 & 0.98   & 0.98 & 0\\
\hline
Ours(w/o GPS)            & \textbf{0.44} & 0.8  & \textbf{1.06}   & \textbf{0.34} & 0.72 & \textbf{1.08} & 0.57  & \textbf{0.62}  & \textbf{0.3} & \textbf{0.32} \\
Ours(w GPS)              & \textbf{0.42} & \textbf{0.66} & \textbf{1.16}   & \textbf{0.67} & \textbf{0.49} & \textbf{1.01} & \textbf{0.52}  & \textbf{0.89 }   & \textbf{0.5} & 0.39 \\
\bottomrule
\end{tabular}
}
\end{sc}
\end{small}
\end{center}
% \vskip -0.26in
\end{table*}

\section{Empirical Evaluation}
%\vskip -0.15in
We conduct comprehensive experiments on two real-world ICU patient datasets to assess our method.
\begin{table*}[t]
\caption{Test set performance of prior work and ours policies.}
%\vskip -0.15in
\label{rlvsours}
\begin{center}
\begin{small}
\begin{sc}
\resizebox{\textwidth}{!}{
\begin{tabular}{l|cccc|cccc}
\toprule
& \multicolumn{4}{c|}{MIMIC-IV} & \multicolumn{4}{c}{HIRID} \\
Policy & $\Delta X \uparrow$ & Cost$\downarrow$ & $\mathcal{L}_b^{test} \downarrow$ & info gain$\uparrow$ & $\Delta X \uparrow$ & Cost$\downarrow$ & $\mathcal{L}_b^{test} \downarrow$ & info gain$\uparrow$\\
\hline
%\midrule
Physician & 0.41 & 0.67 & 1.24 & 0.99   & 0.96 & 0.55 & 0.98 & 1.03\\
\hline
RL\citep{chang2019dynamic}(low cost)& - & \textbf{0.62} & 1.8 & 1 & - & \textbf{0.51} & 1.4 & 1 \\ 
RL\citep{chang2019dynamic}(high cost)& - & 0.8 & 2.3 & \textbf{1.4} & - & 0.74 & 3.6 & \textbf{1.19}\\ 
Ours(w/o GPS) & \textbf{0.44} & 0.8 & \textbf{1.06} & \textbf{(1.2)} & \textbf{1.08}  & 0.57   & \textbf{0.62} & \textbf{(1.17)} \\
Ours(w GPS)    & \textbf{0.42} & \textbf{0.66} & \textbf{1.16} & (0.98) & \textbf{1.01}  & \textbf{0.52}  & \textbf{0.89} & \textbf{(1.05)}\\
\bottomrule
\end{tabular}
}
\end{sc}
\end{small}
\end{center}
% \vskip -0.3in
\end{table*}
%\vskip -0.1in
\textbf{MIMIC and HIRID:} The MIMIC-IV database~\citep{johnson2023mimic} contains anonymized health records from patients who stayed in the ICU of a U.S. hospital. 
The HiRID~\citep{hyland2020early} is a publicly available critical care dataset that includes high-resolution data from patients admitted to an ICU in Switzerland.
Our aim is to develop an optimal policy for daily lab test ordering in ICU patients, maximizing the utility of each test order ($Y$). In this context, every lab test order ($T$) is conceptualized as a $K$-dimensional binary vector. Based on clinical recommendations, we focus on $K=10$ routinely conducted blood tests. 
We analyzed $n=57,212$ valid patient ICU stays for MIMIC-IV, each characterized by $71$ irregular time-series features and $n=32,216$ patient stays from HiRID, each with $73$ features \footnote{The EHR systems for both datasets differ, resulting in a variation in the number of features considered.} mirroring clinician daily practices.
These features encompass lab test results, vital signs, and patient treatments.
Further details on the preprocessing of the MIMIC-IV and HiRID are provided in Appendix~\ref{apx:data}.
%\vskip -0.1in

% \textbf{Patient Status Forecasting:} 
\subsection{Patient Status Forecasting}
We train a multivariate time-series forecasting model based on observed ICU stays to obtain predicted future patient statuses. The dataset, denoted as $\mathcal{D} = \{x^i_{prev}, x^{i*}_{post}\}_{i=1}^n$, consists of $x^i_{prev} \in \mathbb{R}^{48 \times 71}$ representing the patient's past 48-hour ICU stay, and $x^{i*}_{prev} \in \mathbb{R}^{24 \times 71}$ reflecting the \textit{observed} true patient status for the subsequent day. We choose PatchTST \citep{nie2022time} for our patient status forecasting model $\phi$. %due to its superior performance compared to other methods.%(see Table~\ref{sample-table}). 
The training of $\phi$ focuses on minimizing the mean squared error between the predicted future status $x^i_{post}$ and observed future \footnote{Additional details on the training and testing of our forecasting model are available in Appendix~\ref{apx:forecast}.}.
%\vskip -0.1in

% \textbf{Policy Training and Evaluation:} 
\subsection{Policy Training and Evaluation}
Our dataset was partitioned into training, validation, and test sets at proportions of $70\%$, $10\%$, and $20\%$ respectively. Initially, we utilize Algorithm~\ref{alg:orderbound} to establish order bounds for each patient stay. Subsequently, we train our estimated GPS function $\hat{f}(t,x)$, preserving the model parameters that has the lowest validation loss. We then employed Algorithm \ref{alg:bounds}, executing $m=5$ random restarts, to derive $\pi_\theta^{\text{rel}} : \mathbb{R}^{72\times 71} \rightarrow [0,1]^{10}$ that yields the highest outcome on the GPS-constrained validation set. We then evaluate the policy with the best validation set performance on test set with our outcome function $g(t,x)$. Further training specifics, including hyperparameters, are detailed in Appendix \ref{apx:cnf} \& \ref{apx:pihyper}.
%For evaluation, we slightly modified our lab test order utility function $g(t,x)$ to suit as a metric. During testing, we set $\beta_1 = \beta_2 = 1$ for outcome calculation in Eq.~\eqref{eq:gtx}. Additionally, we adapted the smooth term $\mathcal{L}_b$ from Eq.~\eqref{eq:smoothbound} to a discrete form: $\mathcal{L}^{test}_b(t,x) = L_{low} + L_{up}$, where $ L_{low} =\sum_{j=1}^K \mathbbm{1}(t_j < 0.5, t_j^{low} = 1)$ and $L_{up} =\sum_{j=1}^K \mathbbm{1}(t_j > 0.5, t_j^{up} = 0)$.
%$L_{up}$ indicates the redundant tests ordered, and $L_{low}$ represents the essential tests missed by $t$. $\Delta X$ quantifies the variability (information) of the clinician's test order $t$, while $C(t,x)$ denotes the actual lab test cost.
%\vskip -0.1in
\subsection{Results}

\textbf{Our model exceeds all baseline methods.}
To date, no studies have directly focused on deployment-guided off-policy learning for ICU lab test ordering. However, as each test order $t$ is a binary vector, we compared our trained policies against random, lower and upper bound policies, and, crucially, the observed clinician policy (Table~\ref{baselinepolicies}). 
%In Table~\ref{baselinepolicies}, we evaluate our baselines including random policies with 50\% and 75\% chance ordering a test. 
With increasing orders and costs, the information $\Delta X$ provided by these policies also rises. Nevertheless, random policies, despite higher costs, yield no substantial information. From lower bound to physician, and then to upper bound policy, we observe a trend of increasing test order information and cost. Since bound policies represent the limits of our test order space, they exhibit a bound metric of $0$. The physician policy, not entirely aligned with clinical rule-generated orders, incurs some $L_{lower}$. Our learning algorithm is able to discover policies with higher $\Delta X$ and lower out-of-bound test orders $\mathcal{L}_b$ at minimal cost. The ability to consistently find a policy that reduces costs, provides more information, and adheres more closely to clinical guidelines compared to physician policies in historical data highlights the promise of our method as a potential first clinically deployed clinician-facing tool for ICU lab test ordering.
%\vskip -0.1in

% \textbf{GPS for Reliable Policy Learning:} 
\textbf{Our GPS approach for reliable policy learning outperforms non-GPS methods and the Physician policy.}
Our methodology ensures the discovery of reliable policies based on our causal assumptions and the estimated GPS function. 
As shown in Table~\ref{baselinepolicies}, the average total outcome of a reliable policy is approximately $12\%$ higher than that of a policy trained without GPS constraints. Notably, both our reliable and naively trained policies surpass the Physician policy by maintaining lab test orders within bounds or reducing costs, all while providing high information value to clinicians.
%We set the threshold $\overline{\varepsilon}$ at the 5\%-quantile of all $\hat{f}(x,t)$ in the training dataset. 

%\vskip -0.1in
% \textbf{Comparing with RL Policy:} 
\textbf{EXOSITO matches RL approaches in information gain while reducing ordering costs and minimizing out-of-bound orders.}
Although our approach develops a clinician-facing tool within causal contextual bandits, rather than a patient-facing setup based on RL settings where `next state' is observed, 
we find it necessary to compare our results with the work of \citet{chang2019dynamic}. 
While their methodology treats patient status as an hidden state vector of a mortality classifier, making the evaluation of $\Delta X$ intractable, we can still compare policies based on $\mathcal{L}_b$ and cost. We observed that a reward system solely based on a single value (mortality) is inadequate for the lab test ordering problem, as the learned policy does not directly benefit the patient. In Table~\ref{rlvsours}, RL policies derived from \citet{chang2019dynamic}'s method tend to have either low cost with a high $\mathcal{L}_b$ or high cost with significant out-of-bound orders. Interestingly, under their policy evaluation framework, which calculates cumulative information gain from the mortality classifier, our methods demonstrate comparable performance without given any mortality information. This further underscores that mortality is an unsuitable reward metric for lab test ordering.

%Their approach conceptualizes measurement scheduling as a deep Q-learning task in an offline-RL setting. 
% We adapted their method to our dataset, treating each lab test order as an action. 

%\vskip -0.1in
% \textbf{Policy Explainability and Rule Learning:} 
\textbf{Rule integration enables learning transparent, clinically grounded policies.}
Figure~\ref{fig:rules} demonstrates the explainability of our policy, illustrating how lab test orders are linked to both past observations and future predictions of patient status, with each recommendation supported by a patient status time-series matrix for clear clinical rationale. Our experiments further validate the policy's ability to adhere to basic clinical guidelines by showing that adjustments in the Blood Transfusion variable or extreme changes in predicted lab values lead to an increase in specific test orders. These findings, detailed in Appendix~\ref{apx:exp}, underscore the policy's capacity to integrate critical clinical insights, enhancing its applicability and trustworthiness in a healthcare setting.
%\vskip -2in

\textbf{Integrating forecasting, clinical rules, and cost is crucial for robust and minimal lab ordering policies.} Our reward function consists of three components that depend on the performance of our learned patient status forecasting time-series model ($\Delta X$), clinical rules as privileged information ($\mathcal{L}_b$), and the relative clinical cost $C(t)$. We perform a detailed set of ablation studies, including ablating elements of the outcome function, varying the accuracy of the patient forecasting model, changing the conservativeness of the clinical rules, and switching between real-relative cost and uniform cost (see Appendix \ref{apx:x_future} to \ref{apx:rule_ablation}). We found that each element of our outcome function is crucial for policy learning and that our learned policy benefits from including patient status forecasting, clinical rules, and real-relative costs. Moreover, our time-series model performs solidly in predicting patient status, as evaluating our learned policy with perfect predictions yields results that are not significantly different from those using our learned model's predictions. Our ablation study also shows that having rules help the policy generate minimal lab orders and having less conservative rules (i.e. more 1's in the $t^{lower}$) would revert the policy back to the logging policy which errs on over-ordering.

\section{Discussion and Future Work}

In this paper, we introduce, ExOSITO, a novel approach for learning optimal lab test ordering policies for ICU patients, focusing on reliability and explainability. Our method addresses the limited overlap in treatment and covariate spaces. 
We also demonstrate how to leverage privileged information in the form of domain expertise from clinicians to improve the process of learning causal bandits for ICU lab test ordering. Nonetheless, challenges such as ignorability may persist due to potential unobserved confounders and the presence of noisy or corrupted data. Although the EHR dataset encompasses a comprehensive range of clinically relevant variables, unobserved confounding is inevitable given the complexity of the ICU environment. We plan to incorporate as many clinically verified covariates as possible into our policy learning framework to mitigate these issues.

A critical element of our approach is the accurate forecasting of patient future status.
Future improvements might include integrating graph-based time-series models to enhance the accuracy and interpretability of patient status predictions. Alongside structured knowledge, the role of missingness in observed irregular time-series patient data is pivotal for forecasting accuracy, prompting us to assume our data is Missing At Random (MAR) to mitigate its effect. 

Off-policy Evaluation (OPE) is not used in our setting for two primary reasons. First, our dataset does not include `true' outcome measures that quantify the utility of lab test orders, which are essential for mapping state-action pairs to clinical outcomes. Second, the logging policy that generated our data is suboptimal and fails to cover the full range of lab test orders, resulting in insufficient support in action ($T$) space. Together, these limitations prevent OPE from providing a reliable evaluation of our learned policy.

The trained policies of study does not account for potential distribution shifts. 
When applying these policies to different patient cohorts, there is a risk of catastrophic forgetting. Therefore, developing methods that can adapt to distribution shifts is crucial.

By reframing the problem as a causal bandit and demonstrating its effectiveness relative to prior methods on real-patient datasets, our work establishes a robust foundation for subsequent later empirical studies.
Our methodology paves the way for deploying reliable and explainable policies in real clinical settings, with the potential for real-time feedback from medical professionals. With our forecasting setup, we are able to modify the selected features based on the clinicians need. Data collected from such deployments would be invaluable for refining policy accuracy through ground truth labeling and counterfactual learning.

While our results outperform physician orders, we acknowledge that our formulation relies on strong assumptions—namely temporal independence of test ordering decisions and the absence of unobserved confounders—and we will explore semi‐Markov decision processes and augment Equation \ref{eq:smoothbound} with additional safety terms to relax these constraints and enhance robustness. We also plan to rigorously validate our informativeness metric and extend our cost modeling to capture dynamic factors such as test urgency and resource constraints, ensuring clinically important tests are appropriately valued. Finally, to assess real‐world performance and deployment readiness, we will conduct a silent trial in ICU settings to gather direct clinician feedback, compare against rule‐based decision support systems, and systematically analyze potential failure modes of our framework.

%We adopted a general forecasting method PatchTST~\citep{nie2022time}, but ICU data often exhibits strong correlations and underlying structures that could be better captured using a medical knowledge graph. 
%Our current framework utilizes a general off-policy learning and causal setup. However, the outcome function $g(t,x)$ could also serve as a reward function in an offline reinforcement learning (RL) context, provided state and status representations are adequately defined. We demonstrate that using mortality as a sole guide is insufficient for lab test ordering, but offline RL remains a viable avenue for exploration in this domain.

% \newpage 

\bibliography{chil-sample}

\newpage

\appendix

\begin{figure*}[ht]
\begin{center}
\centerline{\includegraphics[width=\textwidth, height=6.3cm]{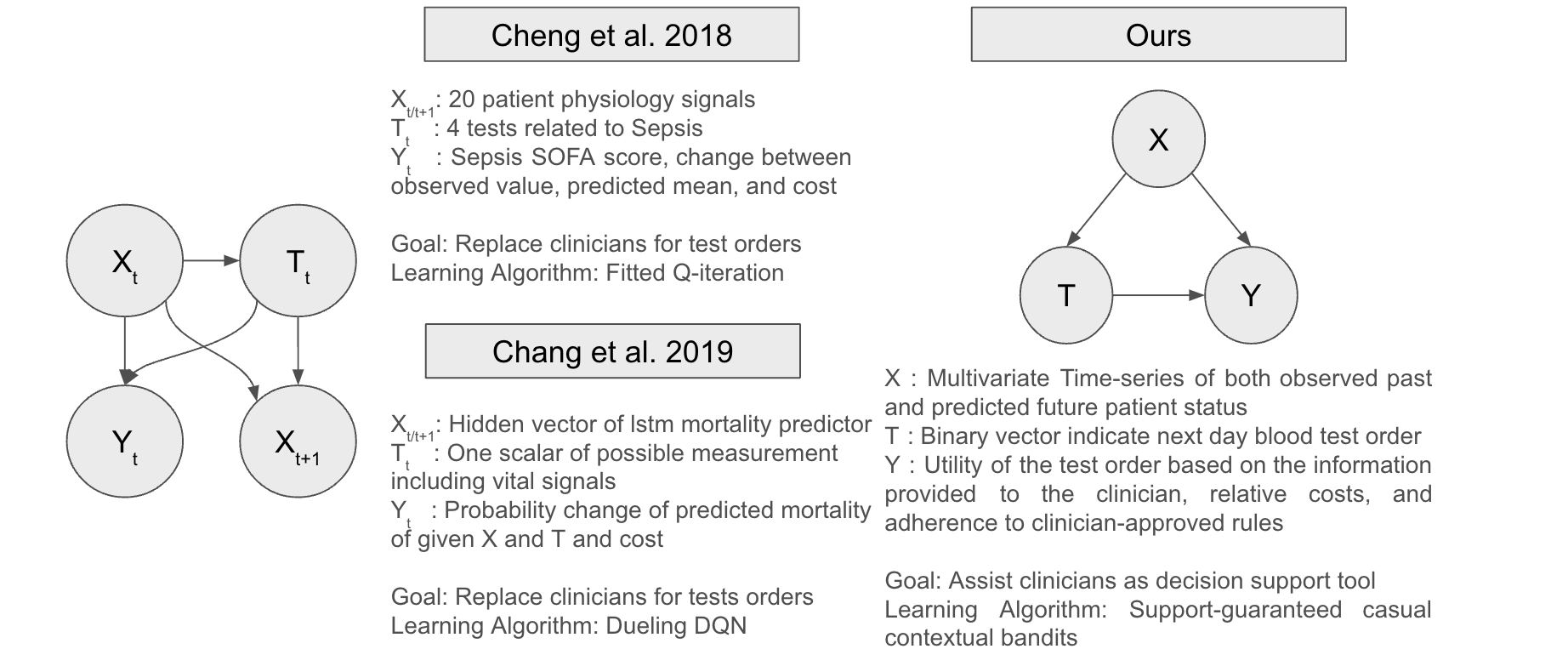}}
\caption{Graphical representation of two prior works~\citep{cheng2018optimal, chang2019dynamic} and our proposed methods.}
\label{fig:graphs}
\end{center}
\vskip -0.4in
\end{figure*}

\section{Datasets and Data preprocessing} \label{apx:data}
MIMIC (Medical Information Mart for Intensive Care) is a publicly available database of de-identified electronic health records (EHRs) from patients admitted to the Beth Israel Deaconess Medical Center (BIDMC) in Boston, Massachusetts. MIMIC-IV~\cite{johnson2023mimic} is one of the largest and most comprehensive critical care databases available, containing data from over 300,000 hospital admissions between 2008 and 2019.

The MIMIC-IV dataset includes a wide range of clinical data, such as vital signs, laboratory test results, medication orders, procedures, diagnoses, and demographic information. The data is collected from various sources, including bedside monitors, electronic medical records, and nursing notes, among others. The data is stored in a relational database format, with each record corresponding to a specific patient encounter.   
To ensure patient privacy and confidentiality, the MIMIC-IV dataset is de-identified and follows the Health Insurance Portability and Accountability Act (HIPPA). It is released under a data use agreement, which requires users to follow strict guidelines for data security and ethical use. However, access to the dataset is free for researchers and clinicians who agree to these terms. 
Overall, the MIMIC-IV dataset is a valuable resource for developing and testing predictive models, evaluating interventions, and improving ICU patient outcomes. With the success of its predecessor, the MIMIC-IV dataset was just released and has not been fully explored.

\textbf{Preprocessing Pipeline for MIMIC-IV dataset}

We develop a set of Python scripts that preprocess and aggregate the MIMIC-IV raw data from relational database format into a format that can be utilized by deep-learning community.  
For developing our preprocessing procedure, we followed and extended a prior work bench-marking the MIMIC-III \cite{johnson2016mimic} with Python~\cite{Harutyunyan2019}. 

We first create a folder indexed by patient subject identification number and extract each patient's raw admission, ICU stays, diagnoses, and laboratory, input/output events information and saved into each patient folder. We then validate the extracted value and unify the missing values obtained from the raw data for each patient. After this step, we prepare the patient ICU stay into time-series data with episodes by event time stamps and store each episode's outcome (mortality, length of stay, diagnoses) in a separate file.  
To reproduce the work done by \cite{chang2019dynamic}, we also generate a script to convert each patient's diagnoses codes into a multi-hot time-invariant features.  

Sitting down with clinical experts in ICU department, we hand-picked the relevant features that clinicians would consider during their daily practice. The features that we considered were Hemoglobin, White cells/White blood cell count, Platelets, Sodium (Na), Potassium (K), Calcium, Phosphate, Magnesium, INR (PT/INR), Alkaline phosphatase (ALP), Bilirubin, ALT, Lactate, Partial pressure of carbon dioxide/PaCO2, PaO2, pH, Bicarb/Bicarbonate, Blood urea nitrogen, Creatinine (blood), Troponin, Creatinine phosphokinase (kinase), Diastolic blood pressure, Mean blood pressure, Systolic blood pressure, Temperature, Heart Rate, Arterial Blood Pressure mean (ABPm), Urine Output, Fluid balance, Fraction inspired oxygen (FiO2), RR Respiratory Rate, Ventilation (mode) Ventilation Mode, PEEP Positive End-Expiratory Pressure, Vt Tidal Volume, GCS Glasgow Coma Scale, SAS Richmond Agitation-Sedation (RAS) Scale, ICDSC Intensive Care Delirium Screening Checklist, Sedation (infusions), Analgesia (infusions), Antipsychotics, Dialysis (yes/no), Vasopressors (IV/PO), Dialysis (output), TPN, Transfusions of blood products, liver toxic drug, Antibiotics, Prone position, NO, Paralysis, Steroids, Diuretics, Antihypertensives (IV/PO), Anticoagulants, Antiepileptics, Enteral nutrition, PPI, Antiarrhythmics, Xray, US, MRI, CT scans, EKG, EEG, ECHO, Hepatotoxic drugs. However, MIMIC-IV data doesn't have any occurrences or record of certain features (e.g. Ultra Sound or NO), we finally picked 71 features with some merging of features with different code like Temperature (\degree F) and Temperature (\degree C) and some non-merged feature like Vasopressors. We show the occurrences of features we considered in Table~\ref{tab:occurs}.

Finally, we convert each patient stay episodes into a patient status forecasting dataset $\mathcal{D} = \{X_i\}_i^N$ where $X$ represents a irregular time-series matrix of patient stay $i$. Among the features in Table~\ref{tab:occurs}, the first 21 features are the test result values correlated with 10 common blood test we consider to order/not order in this paper.

The labels of the dataset are indicators of whether the patient passed away after their ICU stay.  
In order to perform our irregular time-series patient mortality classification, we have to check whether each ICU stay's end time is before the record time of death of the patients. In order for our model to learn meaningful representation, we also eliminated the ICU stays with duration less than 12 hours and stays that has less than 5 lab tests ordered.

After preprocessing with these basic criterion of the ICU stays, we selected 57,212 ICU stays. The morality rate of the total stays is around 12\%. 

\subsection{Discussion of Ignorability Assumption} \label{apx:ignore_assumption}

To uphold our ignorability assumption, we consider our covariate set, which includes treatments (for example, drugs and procedures), vital signals, and test results, to be complete. In other words, any drugs given in the past but not captured in these covariates are assumed to have no effect on lab test orders. If our covariates were to miss essential information for determining lab test orders, the ignorability assumption would no longer hold because unmeasured confounders would be present.

When might this assumption fail? In cases where a drug’s effect is both intense at a single time point and expected to extend over multiple future steps, the bandit framework may not fully capture its impact. This naturally leads us to consider strategies for relaxing this assumption.

One approach is to explicitly relax the assumption by incorporating the dynamic effects of patient covariates—including drug prescriptions—on lab test ordering, effectively framing the problem as an offline reinforcement learning task within a Markov Decision Process. While this extension could broaden EXOSITO's application to various diseases, we believe the bandit model is well-suited for the ICU context, where patient biomarkers are monitored and treatments are frequently adjusted. Alternatively, a simpler solution is to adjust the covariates so that the extended effects of drugs are represented over time while still using the bandit framework, thereby ensuring that long-lasting drug influences are considered when determining lab test orders.

\begin{table*}[ht]
\centering
\caption{Occurrences of selected feature on MIMIC-IV dataset}
\label{tab:occurs}
\begin{tabular}{lc|lc}
\toprule
Measurement & Occurrence & Measurement & Occurrence \\
\midrule
Hemoglobin & 272244 & Heart Rate & 4595306 \\
WBC & 260106 & Urine Output & 2381884 \\
Platelets & 261397 & Respiratory Rate & 4549152 \\
Sodium & 289172 & Ventilation & 443722 \\
Potassium & 360470 & PEEP & 433118 \\
Calcium & 267206 & Tidal Volume & 411535 \\
Phosphate & 265319 & GCS & 3468710 \\
Magnesium & 287194 & SAS & 208084 \\
INR & 183070 & ICDSC & 207757 \\
ALP & 72296 & Sedation & 416082 \\
Bilirubin & 73361 & Propofol & 281361 \\
ALT & 73237 & Analgesia & 346169 \\
Lactate & 156908 & Antipsychotics & 9472 \\
PaCO2 & 256789 & Dialysis & 242060 \\
PaO2 & 218941 & Vasopressors & 439452 \\
ph & 258136 & TPN & 5655 \\
Bicarbonate & 285777 & Transfusions & 54684 \\
Creatinine & 291750 & Prone Position & 2859 \\
Blood Urea Nitrogen & 295945 & Paralysis & 12179 \\
Troponin & 28772 & Diuretics & 81959 \\
Creatinine Kinase & 35863 & Antihypertensives & 199045 \\
Diastolic Blood Pressure & 4504384 & Anticoagulants & 88757 \\
Mean Blood Pressure & 4507939 & Antiepileptics & 18394 \\
Systolic Blood Pressure & 4510870 & Enteral Nutrition & 8939 \\
Temperature & 1068275 & PPI & 89363 \\
FiO2 & 571814 & Antiarrhythmics & 14216 \\
\bottomrule
\end{tabular}
\end{table*}

\section{Details of Building Multivariate Time-Series Patient Status Forecasting Model} \label{apx:forecast}
\begin{table}[t]
%\vskip -0.2in
\caption{Test set performance for trained time-series forecasting models. We found that the PatchTST model obtained the lowest forecasting error and formed the backbone for our forecasting module}
\label{sample-table}
\vskip -0.18in
\begin{center}
\begin{small}
\begin{sc}
\begin{tabular}{lcccr}
\toprule
Model & MSE & MAE \\
\midrule
Linear    & 0.037$\pm$ 0.002& 0.094$\pm$ 0.005 \\
LSTM   & 0.035$\pm$ 0.004& 0.072$\pm$ 0.002\\
PatchTSMixer  & 0.032$\pm$ 0.066& 0.066$\pm$ 0.003       \\
PatchTST    & \textbf{0.027$\pm$ 0.001}& \textbf{0.059$\pm$ 0.002}\\
\bottomrule
\end{tabular}
\end{sc}
\end{small}
\end{center}
\vskip -0.35in
\end{table}
In our study, we developed a multivariate time-series model to forecast patient status, leveraging deep learning techniques for both short and long-term predictions. We evaluated various architectures including simple linear transformations, Long Short-Term Memory (LSTM) networks~\cite{hochreiter1997long}, PatchTSMixer~\cite{ekambaram2023tsmixer}, and PatchTST~\cite{nie2022time}, focusing on their ability to accurately predict future patient states based on historical data.

PatchTST stands out for its innovative approach to handling time-series data, treating inputs and outputs as matrices to effectively process information across multiple variables. By dividing the input matrix $x \in \mathbb{R}^{d \times L_1}$ into subsequences or 'patches,' PatchTST captures temporal dynamics with precision. These patches undergo processing through transformer blocks, adept at modeling dependencies along the axes of time and feature dimensionality. Key to PatchTST's architecture are its embedding layer, which elevates the dimensionality of input patches for subsequent processing, and its transformer encoder layers, featuring multi-head self-attention mechanisms and position-wise feed-forward networks. These components enable the modeling of intricate temporal relationships, culminating in an output linearly projected to dimensions $x \in \mathbb{R}^{d \times L_2}$. In our model, we set $L_1 = 48$ and $L_2 = 24$ to predict the next day's patient status using data from the prior 48 hours, employing mean imputation to address irregularities in time-series data.

Training PatchTST necessitates selecting an appropriate loss function, optimizer, and constructing a training regimen. We utilize the Mean Squared Error (MSE) loss for its aptitude in regression tasks, specifically in gauging the accuracy of time-series forecasts. Adam optimizers were chosen for their efficiency with sparse gradients and adaptive learning rates, tested across various initial learning rates ($5e-3, 1e-3, 1e-4, 5e-4$). Additionally, the implementation of a OneCycle learning rate scheduler alongside three other scheduling functions further refines our training process.

For LSTM configurations, we opted for a three-layer setup with 512 hidden dimensions, adhering to standard configurations for both PatchTST and PatchTSMixer to ensure consistency in model evaluation. The effectiveness of these models was determined based on MSE loss performance on a validation set, with comparative results detailed in Table~\ref{sample-table}.

\subsection{Evaluating Patient Status Forecasting Model}

As mentioned in the discussion session, the performance of the time-series forecasting model can impact the subsequently learned policy. To better clarify this point, we conducted a new evaluation of our time-series predictor on the test set, assessing performance (MSE) by calculating the difference at each time point (each hour of the predicted future 24 hours) and for each feature. Our analysis verified that across all time points, the MSE consistently hovers around 0.03, which aligns with our reported results.

To gain further insights, we evaluated the MSE for each feature across all test set samples and grouped features into three categories based on their average MSE:

\begin{itemize}
    \item Low MSE Group
    
Features (Counts):
CreatinineKinase (6977), MRI (931), PronePosition (638), Vasopressors (86261), Antipsychotics (1924), TPN (1283), UltraSound (1555), EnteralNutrition (1844), CTScan (2998), Antiarrhythmics (2705), PReplacement (2199), Paralysis (2591), Antiepileptics (4210), ICPMonitor (21547), Troponin (5656), SAS (40291), Calcium (53351), Lactate (32035), MgReplacement (10866), Sodium (58406), Magnesium (57475)

Average MSE: 0.0059

    \item Medium MSE Group

Features (Counts):
TidalVolume (79486), Hemoglobin (54710), PaCO2 (51837), Bicarbonate (57018), Xray (12783), ALP (14574), Transfusions (10977), FiO2 (115116), ph (52246), ICDSC (32044), WBC (52127), Bilirubin (14774), Potassium (72510), AirwayPressure (80863), INR (36491), ALT (14720), Anticoagulants (18484), Phosphate (53024), Diuretics (16520), PPI (18118), PEEP (86456)

Average MSE: 0.0195
    \item High MSE Group
    
Features (Counts):
Platelets (52386), CaReplacement (13279), UrineOutput (478588), Antihypertensives (38794), MinuteVentilation (81693), BloodUreaNitrogen (59222), Dialysis (30275), Temperature (213724), PaO2 (44297), Creatinine (58271), GCS (227697), KReplacement (31494), Analgesia (70894), Antibiotics (70132), Sedation (83964), DiastolicBloodPressure (881326), MeanBloodPressure (881395), HeartRate (917866), SystolicBloodPressure (882387), RespiratoryRate (909106), Ventilation (88295)

Average MSE: 0.0801

\end{itemize}

We observed that vital signals (e.g., heart rate), which require less manual effort to collect, are more frequently represented in the data, whereas treatments and lab values are more sparsely recorded, reflecting the real-time nature of patient monitoring. Despite these challenges, we believe incorporating the forecasting model into our approach enhances the ability of the learned policy to make clinically relevant lab order recommendations based on both present and (predicted) future for patient status.

\section{Detailed Rules for Necessary Lab Test Orders} \label{apx:boundgen}

In our study, we focus on ten blood tests frequently ordered in clinical settings: Complete Blood Count (CBC), Electrolytes, Calcium Profile, INR, Liver Profile, Lactate, Arterial Blood Gas (ABG), Creatinine, Troponin, and Creatinine Kinase (CK). We derived a set of lower bound rules for these test orders after extensive discussions with medical experts who are senior attending physicians in leading hospitals together with support form peer-reviewed literature~\citep{kumwilaisak2008effect, cismondi2013reducing, vidyarthi2015changing, bindraban2018reducing}, which are encapsulated in the rule set $\mathcal{CR}$ used in Algorithm~\ref{alg:orderbound}:

\begin{itemize}
    \item If patient receives blood transfusion, then order CBC, Electrolytes, and INR.
    \item If patient Urine Output of the last 24 hours is less than 1 liter or greater than 4 liters, order Electrolytes and Creatinine.
    \item If patient had $25\%$ increasing dose of Vasopressors (or receiving new Vasopressor), order CBC, Liver Profile, Troponin, Lactate.
    \item If patient had dialysis or will have dialysis, order Calcium Profile.
    \item If patient has a new fever, order CBC and Liver Profile.
    \item If the patient Minute Ventilation is increased or decreased by 25\%, order ABG.
    \item If the patient Airway Pressure has 25\% increase, then order ABG.
    \item If the patient had Antibotics treatment, order CBC.
    \item If the patient had Antiarrhythmics treatment, order Calcium Profile and Electrolytes.
    \item If the patient had Anticoagulants treatment, order INR.
    \item If the patient had Propofol treatment, order CK.
    \item If the patient is on ICP Monitor, order Electrolytes.
    \item If the patient White Blood Cell (WBC) is less than 1 or greater than 12, order CBC and Liver Profile.
    \item If the patient White Blood Cell (WBC) has 5 unit of change in the past 24 hours, order CBC.
    \item If Creatinine value greater than 150 or has 50 increase in the past 24-48 hours, order ABG, Electrolytes, and Calcium Profile.
    \item If the patient Creatinine Kinase value greater than 5000, order CK.
    \item If the patient PEEP value has increase more than 2, order ABG.
    \item If the patient PH is less than 7.3, order Lactate and Creatinine.
    \item If the patient Hemoglobin value is less than 7, order CBC and INR.
    \item If the patient Hemoglobin value has decreased more than 2 unit in the past 24 hours, order CBC.
    \item If patient Platelets is less than 30 or greater than 600000, order CBC.
    \item If patient Platelets value has more than 30\% decrease in the past 48 hours, then order CBC.
    \item If patient had K replacement in the past 12 hours, order Electrolytes.
    \item If patient had Ca replacement in the past 12 hours, order Electrolytes.
    \item If patient had P replacement in the past 12 hours, order Electrolytes.
    \item If patient had Mg replacement in the past 12 hours, order Electrolytes.
    \item If patient Sodium (Na) has 6 unit change in the past 24 hours, order Electrolytes.
    \item If patient Sodium (Na) is greater than 150 or less than 135, order Electrolytes.
    \item If patient Potassium (K) is greater than 5 or less than 3.5 order Electrolytes.
    \item If patient Potassium (K) is greater than 4.5, order Creatinine.
    \item If patient Calcium is greater than 3 or less than 2, order Calcium Profile.
    \item If patient Phosphate is greater than 0.6 or greater than 1.8, order Calcium Profile.
    \item If patient Magnesium is greater than 0.8, order Calcium Profile.
    \item If patient INR is greater than 1.6, order INR.
    \item If patient Alanine Transaminase (ALT) is greater than 100, order liver profile.
    \item If patient Bilirubin is greater than 50, oreder liver profile.
    \item If patient uses Hepatotoxic drug, order liver profile.
    \item If patient has Arrhythmia, order Troponin and Calcium Profile.
    \item If patient had Diuretics, order Calcium Profile.
\end{itemize}

Despite the comprehensive suite of rules applied to our patient ICU stay dataset, it's important to note that these rules were crafted with a high degree of conservatism. This approach is in alignment with clinical practices, ensuring that the rules are seen as necessary and reasonable by healthcare professionals. For each patient stay $x$ in our dataset, we utilize Algorithm~\ref{alg:orderbound} to generate a binary vector of length 10, indicating the ordered tests for the following day.

\begin{figure*}[ht]
\begin{center}
\centerline{\includegraphics[width=\textwidth]{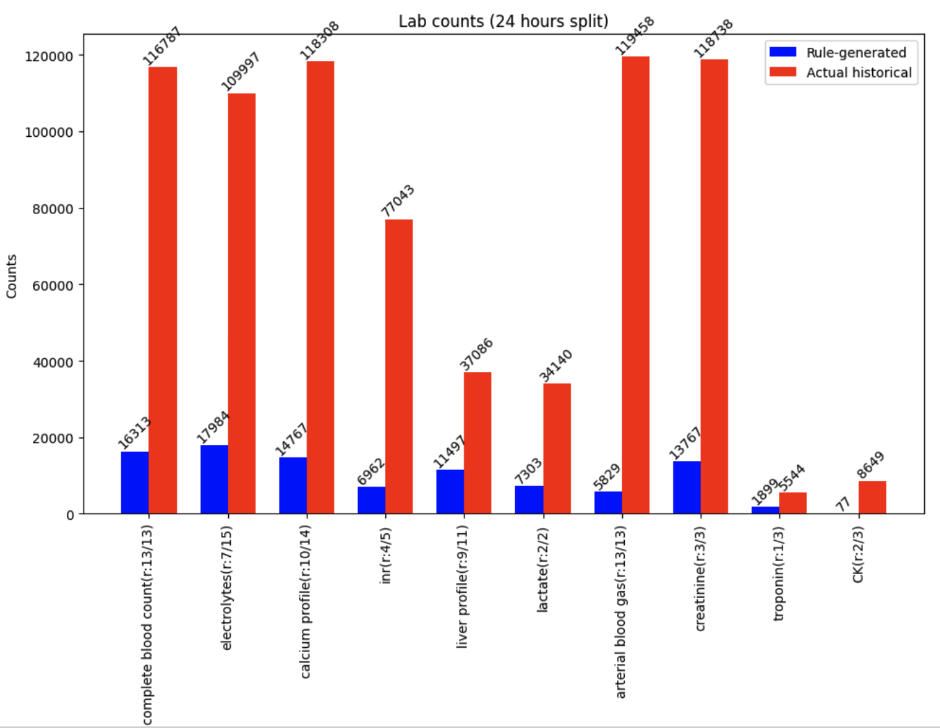}}
\caption{Bar plot that shows the distribution of guideline-generated tests and observed tests}
\label{fig:rulevsobs}
\end{center}
\vskip -0.4in
\end{figure*}

In Figure~\ref{fig:rulevsobs}, we present a comparison between the orders generated by our rules and the orders actually placed by physicians. This visualization serves to highlight the extent to which our algorithmically generated orders align with real-world clinical decision-making.

\section{Design of $\Delta X$} \label{apx:deltax}

During consultations with practitioners, clinicians noted that most patients tend to have consistent lab results, which are often classified as ‘abnormal.’ However, our clinical collaborators emphasized that it is more informative to observe extreme or mean changes in lab results. The motivation for designing $\Delta X$ to measure the range or mean changes is that clinicians are more likely to order a test if the test result is expected to change significantly (either improving or deteriorating).

As introduced in Section \ref{deltax}, our framework assumes that lab test orders are more informative to clinicians when there are larger changes in lab values—whether these are extreme changes or significant shifts in mean values. $\Delta X$ is specifically designed to encourage the policy to prioritize tests that are likely to provide such informative insights.

The design choice of $\Delta X$ is supported both by our conversations with clinical collaborators and prior literature.

\citet{ccakirca2023comparing} and \cite{mahmood2023hematological} both provide a comparative analysis of blood test abnormalities in ICU vs. non-ICU COVID-19 patients. The results show that WBC are remarkably higher in the ICU patients than non-ICU patients while albumin levels are remarkably lower for ICU patients compared to non-ICU patients. The authors concluded that ICU patients' conditions are more severe than the general patient population outside the ICU. 

In \citet{xu2021varying}, the authors specifically compared lab results of creatinine, hemoglobin, lactate, sodium and showed that ICU patients normally have a much higher variance in these test results and used reference ranges are larger than non-ICU patients. This study suggests that standard reference ranges have limited relevance for ICU patients and highlights the need for context-specific ranges to enhance clinical interpretation. 

The ICU-Labome study \cite{alkozai2018systematic} evaluated 35 routine laboratory measurements in over 49,000 ICU patients. The research found that many laboratory values were outside standard reference intervals. Notably, 14 out of 35 measurements had median values outside standard reference intervals, underscoring the necessity for ICU-specific reference ranges. 

Authors in \cite{mamandipoor2022prediction} build machine learning models that predict lactate values for ICU patients and show that elevations in serum lactate levels are strong predictors of mortality. Thus, anticipating these changes allows clinicians to intensify care proactively. 
In addition, ordering routine tests for ICU patients is common medical practice. However, studies also suggest that continuing to order certain lab tests when their values remain stable may not be beneficial and can even be harmful \citep{kleinpell2018reducing, eaton2017evidence, levi2019reducing}. 

\citet{kleinpell2018reducing} recommend against routine daily laboratory tests for clinically stable ICU patients, as such practices often do not contribute to patient care and can lead to unnecessary interventions. Excessive blood draws for stable ICU patients can lead to hospital-acquired anemia, increased healthcare costs, and unnecessary downstream testing and procedures \citep{eaton2017evidence}. Implementing strategies to reduce unnecessary laboratory testing in the ICU has been shown to decrease the volume of blood collected per patient-day without negatively affecting patient outcomes \citep{levi2019reducing}. 

The references provided support the understanding that ICU patients generally exhibit more consistently abnormal laboratory values compared to patients in other wards. Building on this, we argue that instead of concentrating on lab values that are stable yet abnormal, it would be more clinically valuable to focus on identifying and prioritizing tests for lab values that demonstrate significant changes over time. These dynamic changes are more likely to provide actionable insights and support timely clinical interventions.

\section{Estimate Propensity Score Function with Conditional Normalizing Flow}  \label{apx:cnf}
This section elaborates on the foundational concepts and specific methodologies employed for developing our approximate generalized propensity score (GPS) function, denoted as $\hat{f}(x,t)$.

Normalizing Flows initially emerged to enhance the variational inference in variational autoencoders, as documented in seminal works \cite{rezende2015variational, tabak2010density}. These models operate by converting a straightforward initial distribution, for instance, a Gaussian, into a complex distribution that closely resembles the distribution of the actual data. This is achieved through a sequence of reversible mappings. We represent the initial distribution by $p_z({z})$, with ${z}$ being a latent variable, and the distribution of the actual data by $p_x({x})$, where ${x}$ indicates the observed data. The objective is to discover a function $f$ that facilitates the transformation ${x} = f({z})$.

At the heart of normalizing flows lies the concept of utilizing a composition of bijective functions $f = f_K \circ f_{K-1} \circ \ldots \circ f_1$, with $K$ indicating the count of these transformations. An affine transformation usually represents the final transformation $f_K$, while preceding transformations are reversible and nonlinear. Due to the invertibility of each function $f_k$, the reverse mapping is straightforwardly computed.

The probability density of ${x}$ in relation to ${z}$ is calculable through the change of variables theorem. Assuming the base distribution $p_z({z})$ is well-defined and simple (e.g., Gaussian), the density of ${x}$ can be determined as follows:
\begin{equation*}
p_x(x) = p_z(z) \left|\det \left(\frac{\partial f}{\partial z}\right)\right|^{-1}
\end{equation*}
Here, $\left|\det \left(\frac{\partial f}{\partial {z}}\right)\right|$ signifies the determinant of the Jacobian matrix of transformation $f$ relative to ${z}$, illustrating the alterations in the latent space density to align with $x$'s density.

For inference tasks like density estimation or sampling, it is crucial to compute the log-likelihood of the observed data ${x}$. Considering a dataset $\mathcal{D} = {{x}_1, \ldots, {x}_n}$, the log-likelihood sums up the log-densities for each data point:
\begin{equation*}
    \log p(\mathcal{D}) = \sum_{i=1}^n \log p_x(x_i) .
\end{equation*}

In our implementation, we adopt conditional normalizing flows (CNFs)~\cite{trippe2018conditional,winkler2019learning} following \cite{schweisthal2023reliable} for the GPS estimation. CNFs adapt the concept of normalizing flows to conditionally model densities $p(y \mid x)$ by applying an invertible transformation to a base density $p(z)$, with transformation parameters $\gamma(x)$ reliant on the input $x$.

Utilizing neural spline flows \cite{durkan2019neural} in conjunction with masked auto-regressive networks \cite{dolatabadi2020invertible}, our methodology enables modeling the conditional distribution of data variables based on a conditioning variable, sequentially generating each variable while considering previously generated variables. This sequential generation underpins efficient computation and sampling from the conditional distribution. CNFs stand out due to their universal approximation capabilities, ensuring accurate density function modeling for complex scenarios, alongside benefits of proper normalization, parametric nature facilitating constant inference time post-training \cite{Melnychuk.2022b}.

For the GPS modeling $\hat{f}(t, x)$, we integrate neural spline flows with masked auto-regressive networks, setting a flow length of $3$, adopting quadratic splines across equally spaced bins. The autoregressive model, a Multilayer Perceptron (MLP) with three hidden layers and 50 neurons each, incorporates noise regularization using noise from $N(0, 0.1)$. The training of CNFs minimizes the negative log-likelihood (NLL) loss, employing the Adam optimizer with a batch size of $512$ over up to 300 epochs, incorporating early stopping based on NLL loss on a validation dataset. Learning rate tuning spans $\{0.0001, 0.0005, 0.001, 0.005, 0.01 \}$, with model evaluation mirroring early stopping criteria. For input handling, $x$, a two-dimensional time-series matrix, is flattened into a vector for processing, ensuring accurate covariate

\section{Policy Learning Algorithm and Hyperparameters for Policy Training }  \label{apx:pihyper}
Integrating the forecasting model, established bounds, and the potential outcome function, we introduce a time-aware, overlap-guaranteed off-policy learning algorithm. This algorithm is designed to create an explainable, reliable, and optimal policy for lab test ordering in ICU environments.

Our objective is to identify a policy $\pi^\text{rel}$ that not only maximizes the estimated policy value $\hat{V}(\pi)$ but also guarantees that $V(\pi)$ is determined \emph{reliably}. To this end, we restrict our policy search to regions within the covariate-treatment domain where data support is substantial, ensuring no violation of overlap. Modifying our original objective from Eq.~\eqref{eq:prob_baseobjective}, we reformulate it as:
\vskip -0.12in
\begin{equation}
    \pi^\text{rel} \in \underset{\pi \in \Pi^{\mathrm{r}}}{\arg\max} \; \hat{V}(\pi)
\end{equation}
\vskip -0.12in
where $\Pi^{\mathrm{r}} = \left\{ \pi \in \Pi\ \,\mid\, f \left(\pi(x), x) > \overline{\varepsilon} \right), \; \forall x \in \mathcal{X} \right\}$ defines our policy class with a reliability threshold $\overline{\varepsilon}$, which dictates the minimum overlap.
Given the constraints of our finite observational data, this leads to the following optimization problem:
\vskip -0.12in
\begin{equation}
%\label{eq:method_constrainedobjective}
    \begin{aligned}
         & \underset{\pi}{\max} & \frac{1}{n}\sum_{i=1}^{n} g \left( \pi(x_i), x_i \right) & &
         \text{s.t.}  & & \hat{f} \left( \pi(x_i), x_i \right) \geq \overline{\varepsilon} &  \\
    \end{aligned}
\end{equation}
\vskip -0.12in
In this framework, the lab test order utility function $g(t,x)$ serves as our policy outcome estimator, and the GPS estimator $\hat{f}(t, x)$ limits the policy search space.
To represent our policy $\pi$, we employ neural networks with learnable parameters $\pi_\theta$. Since the constrained optimization problem in Eq.\eqref{eq:method_constrainedobjective} is not amenable to direct learning through gradient updates, we convert it into an unconstrained Lagrangian problem:
\vskip -0.16in
\begin{equation}%\label{eq:lagrangian_optimization}
    \begin{aligned}
        \underset{\theta}{\min} \; \underset{\lambda_i \geq 0}{\max}  - \frac{1}{n}\sum_{i=1}^{n} \left\{ g \left( \pi_{\theta}(x_i), x_i \right) - \lambda_i \left[ \hat{f} \left( \pi_{\theta}(x_i), x_i \right) - \overline{\varepsilon} \right] \right\},
    \end{aligned}
\end{equation}
\vskip -0.15in
where $\pi_{\theta}(x_i)$ denotes the policy learner with parameters $\theta$, and $\lambda_i$ are the Lagrange multipliers for each sample $i$. This Lagrangian min-max objective is tackled through adversarial learning, employing gradient descent-ascent optimization techniques~\cite{Lin.2020}.

Leveraging our patient status forecasting model $\phi$, the defined outcome estimation function $g$, the estimated GPS function $\hat{f}$, and the min-max-objective in Eq.~\eqref{eq:lagrangian_optimization}, 
we are equipped to establish our explainable and reliable policy, as detailed in Algorithm~\ref{alg:bounds}. 
One important aspect to consider is that despite having defined our outcome estimation function $g$, it is imperative for all operations within $g$ to be differentiable to enable the gradient descent-ascent algorithm to function effectively through backpropagation. In the case of $C$ and $\Delta X$, both employ a step function to ascertain which lab tests are ordered. We address this challenge by employing a modified Sigmoid function to approximate the step function operations.

For our policy network $\pi_\theta$, we opt for a PatchTSMixer architecture. We determine the reliability threshold $\Bar{\varepsilon}$ as the $5\%$-quantile of the estimated GPS $\hat{f}(t, x)$ from the training set, unless specified differently. For the optimization of parameters $\theta$ and $\lambda$, we employ Adam optimizers, considering batch sizes of $\{512, 1024, 2048, 4096\}$. The network is trained leveraging the gradient descent-ascent optimization objective outlined in Eq.~\eqref{eq:lagrangian_optimization}, targeting a maximum of 50 epochs. Early stopping is implemented based on a patience of $7$ epochs for the validation loss, as determined by Algorithm~\ref{alg:bounds} on the factual validation dataset.

The learning rate for updating $\lambda$ is set to $\eta_\lambda= 0.01$. A random search across 10 configurations is conducted to fine-tune the learning rate for updating the policy network's parameters, $\eta_\theta$, within the set $\{5e-3, 1e-3, 5e-4, 1e-4\}$, as well as to initialize the Lagrangian multipliers $\lambda_i$ within the range [1, 5, 10]. Additionally, we explore different values for the outcome function terms, specifically $\beta_1 = \{0, 1, 10, 100\}$ and $\beta_2 = \{0, 1, 10, 100\}$. The performance evaluation during the hyperparameter tuning phase adheres to the same criterion used for early stopping. Subsequent to the hyperparameter determination, we conduct $k=5$ experimental runs to identify the optimal policy setting. Our method is trained with NVIDIA A6000 GPUs, one single A6000 GPU would be able to complete the training.

For evaluation, we slightly modified our lab test order utility function $g(t,x)$ to suit as a metric. During testing, we set $\beta_1 = \beta_2 = 1$ for outcome calculation in Eq.~\eqref{eq:gtx}. Additionally, we adapted the smooth term $\mathcal{L}_b$ from Eq.~\eqref{eq:smoothbound} to a discrete form: $\mathcal{L}^{test}_b(t,x) = L_{low} + L_{up}$, where $ L_{low} =\sum_{j=1}^K \mathbbm{1}(t_j < 0.5, t_j^{low} = 1)$ and $L_{up} =\sum_{j=1}^K \mathbbm{1}(t_j > 0.5, t_j^{up} = 0)$.
$L_{up}$ indicates the redundant tests ordered, and $L_{low}$ represents the essential tests missed by $t$. $\Delta X$ quantifies the variability (information) of the clinician's test order $t$, while $C(t,x)$ denotes the actual lab test cost.

\begin{table*}[t]
\caption{Testset performance for baseline and our learned policies (with std).}
\label{fulltestperf}
\begin{center}
\begin{small}
\begin{sc}

\begin{tabular}{lccccr}
\toprule
Policy & $\Delta X \uparrow$ & Cost$\downarrow$ & $\mathcal{L}_b^{test} \downarrow$ & $L_{low}\downarrow$ & $L_{up}\downarrow$ \\
\midrule
$\text{Random}_{0.5}$    & 0.23 $\pm$0.01 & 0.51$\pm$0.01 & 4.3$\pm$0.01    & 3.2$\pm$0.01 & 1.1$\pm$0.01 \\
$\text{Random}_{0.75}$   & 0.34 $\pm$0.01 & 0.75$\pm$0.01 & 3.64$\pm$0.01   & 1.6$\pm$0.01 & 2.04$\pm$0.01 \\
LowerBound               & 0.37 & 0.62 & 0      & 0 & 0 \\
UpperBound               & 0.44 & 0.82 & 0      & 0 & 0 \\
Physician                & 0.41& 0.67 & 1.24   & 1.24 & 0 \\
\hline
Ours(w/o GPS)            & 0.44 $\pm$0.01& 0.8 $\pm$0.003 & 1.06$\pm$0.001   & 0.34 & 0.72 \\
Ours(w GPS)              & 0.42 $\pm$0.005& 0.66$\pm$0.005 & 1.16$\pm$0.001   & 0.67 & 0.49 \\
\bottomrule
\end{tabular}

\end{sc}
\end{small}
\end{center}
\vskip -0.2in
\end{table*}

\begin{table*}[t]
\caption{Testset performance of prior work and ours policies (with std).}

\label{fullrl}
\begin{center}
\begin{small}
\begin{sc}

\begin{tabular}{lcccr}
\toprule
Policy & $\Delta X \uparrow$ & Cost$\downarrow$ & $\mathcal{L}_b^{test} \downarrow$ & info gain$\uparrow$ \\
\midrule
Physician & 0.41 & 0.67 & 1.24 & 0.99 \\
\hline
RL (low cost)& - & 0.62 & 1.8 & 1 \\ 
RL (high cost)& - & 0.8 & 2.3 & 1.4 \\ 
Ours(w/o GPS) & 0.44$\pm$0.01 & 0.8 $\pm$0.003& 1.06$\pm$0.001 & (1.2) \\
Ours(w GPS)    & 0.42$\pm$0.005 & 0.66$\pm$0.005 & 1.16 $\pm$0.001& (0.98) \\
\bottomrule
\end{tabular}
\end{sc}
\end{small}
\end{center}

\end{table*}

\begin{table*}[t]
\vskip -0.2in
\caption{The test set performance for trained PatchTST time-series forecasting model with different minimum number of ordered tests}
\label{ts-table}
%\vskip 0.15in
\begin{center}
\begin{small}
\begin{sc}
\begin{tabular}{lcccr}
\toprule
Number of minimum test required & MSE & MAE \\
\midrule
10, 5 (Main result)    & 0.027$\pm$ 0.001& 0.059$\pm$ 0.002 \\
5, 3    & 0.024$\pm$ 0.002& 0.054$\pm$ 0.003\\
1, 0  &   0.018$\pm$ 0.002& 0.038$\pm$ 0.001       \\
\bottomrule
\end{tabular}
\end{sc}
\end{small}
\end{center}

\end{table*}

% \newpage
\section{Additional Experiments on MIMIC and HIRID Dataset}

We conducted further experiments on the MIMIC dataset, adjusting for the minimum required lab tests ordered in each data point. These experiments consistently showed that our method, employing GPS-guided policy learning, outperforms those based on physician decisions and conventional RL approaches, with mortality as the reward metric.

In addition to our work on the MIMIC dataset, we aimed to validate the applicability of our method on a broader scale. We analyzed a recently released ICU dataset from patients in a Swiss hospital, known as the High Time Resolution ICU dataset (HiRID)~\cite{hyland2020early}. This dataset, initially utilized to predict circulatory failure, has mostly been explored through its inferred version in previous studies. However, our detailed examination of HiRID’s raw data revealed that it serves as an apt irregular time-series dataset for our objectives, similar in structure to MIMIC. We present an overview of this newly processed HiRID dataset alongside the results of applying our method, mirroring the experimental approach taken with the MIMIC dataset.

\subsection{Various Patient Time-series Covariates Settings}
Our method begins by constructing a time-series forecasting model, $\phi$, that predicts the future ICU stay status of a patient based on observed data. Among the 71 features in our covariates $X$, 21 are related to blood test values. To validate the trained model, our main experiment ensured that for each 24-hour period, at least 5 tests were ordered, meaning $X_{prev}$, covering 48 hours, should include results from at least 10 tests, and $X_{post}$, spanning 24 hours, from at least 5 tests. This requirement explains why the results in Table \ref{fulltestperf} show an average of 6 to 8 tests ordered for the next day.

To prepare for eventual deployment in ICU wards, we tested various settings with even fewer minimum required data points in $X$, aiming to more closely mirror real-world situations where some patients may not undergo any tests within a 24-hour period.

With this consideration, we established two additional settings for the minimum number of required tests: the first requires $X_{prev}$ to include results from at least 5 lab tests and $X_{post}$ from at least 3 tests; the second mandates $X_{prev}$ to contain results from at least 1 lab test, with no requirements for $X_{post}$. Despite these adjustments, we ensured that each data point included at least 15 non-missing entries collected over the patient's 48-hour ICU stay of interest.

As we reduced the minimum number of required tests, we observed an increase in the number of zeros or missing values in the data points. For our PatchTST forecasting model, the loss decreased as the minimum number of required tests was lowered. Given that PatchTST outperformed other models in our forecasting experiments, we evaluated its performance under these three settings of minimum test requirements. The findings are detailed in Table \ref{ts-table}.

Subsequently, we employed these two additional versions of $\phi$ to explore the learning of our lab test ordering policy, guided by our proposed outcome function $g(t,x)$.

\begin{table*}[t]
\caption{Testset performance for baseline and our learned policies (with std). Minimum required tests is 5 per 48 hours.}
\label{min53-tab}
\begin{center}
\begin{small}
\begin{sc}
\begin{tabular}{lccccr}
\toprule
Policy & $\Delta X \uparrow$ & Cost$\downarrow$ & $\mathcal{L}_b^{test} \downarrow$ & $L_{low}\downarrow$ & $L_{up}\downarrow$ \\
\midrule
$\text{Random}_{0.5}$    & 0.21 $\pm$0.01 & 0.51$\pm$0.01 & 4.3$\pm$0.01    & 3.2$\pm$0.01 & 2.1$\pm$0.01 \\
LowerBound               & 0.13 & 0.24 & 0      & 0 & 0 \\
UpperBound               & 0.25 & 0.47 & 0      & 0 & 0 \\
Physician                & 0.17& 0.36 & 1.18   & 1.18 & 0 \\
\hline
Ours(w/o GPS)            & 0.24 $\pm$0.007& 0.46 $\pm$0.003 & 0.94 $\pm$0.003   & 0.13 & 0.81 \\
Ours(w GPS)              & 0.2 $\pm$0.002& 0.32$\pm$0.01 & 1.09$\pm$0.004   & 0.4 & 0.69 \\
\bottomrule
\end{tabular}
\end{sc}
\end{small}
\end{center}
\vskip -0.2in
\end{table*}

\begin{table*}[t]
\caption{Testset performance for baseline and our learned policies (with std). Minimum required tests is 1 per 48 hours.}
\label{min10-tab}
\begin{center}
\begin{small}
\begin{sc}
\begin{tabular}{lccccr}
\toprule
Policy & $\Delta X \uparrow$ & Cost$\downarrow$ & $\mathcal{L}_b^{test} \downarrow$ & $L_{low}\downarrow$ & $L_{up}\downarrow$ \\
\midrule
$\text{Random}_{0.5}$    & 0.21 $\pm$0.01 & 0.51$\pm$0.01 & 4.2$\pm$0.01    & 3.1$\pm$0.01 & 2.1$\pm$0.01 \\
LowerBound               & 0.09 & 0.18 & 0      & 0 & 0 \\
UpperBound               & 0.23 & 0.41 & 0      & 0 & 0 \\
Physician                & 0.15& 0.33 & 1.12   & 1.12 & 0 \\
\hline
Ours(w/o GPS)            & 0.21 $\pm$0.001& 0.38 $\pm$0.002 & 0.91 $\pm$0.003   & 0.07 & 0.84 \\
Ours(w GPS)              & 0.18 $\pm$0.001& 0.28$\pm$0.01 & 1.03$\pm$0.004   & 0.22 & 0.79 \\
\bottomrule
\end{tabular}

\end{sc}
\end{small}
\end{center}
\vskip -0.2in
\end{table*}

\subsection{Consistent MIMIC Results for Different Covariates Settings}

Incorporating two additional settings for the minimum required number of tests, we extended our experiments as detailed in the results section of the paper. Table~\ref{min53-tab} displays the test set performance for both baseline models and our learned policies with a minimum of 5 tests required every 24 hours. Table~\ref{min10-tab} presents the performance for baseline models and our learned policies with only 1 minimum test required every 48 hours.

Firstly, across these settings, we observed a consistent trend: our learned policy, supported by the propensity score function, consistently orders tests that not only provide clinicians with more relevant information but also generate lower costs and adhere more closely to clinical rules, thereby missing fewer necessary tests compared to the physician's policy.

A notable decrease in costs across methods, except for the random policy, was observed. This reduction is attributed to the less frequent testing of patient ICU stays due to the altered minimum test requirements. As evidenced in Table~\ref{min10-tab}, an average of 2 to 3 tests are ordered every 24 hours. Reducing the number of minimally required tests led to an increase in $L_{up}$, indicating a decrease in the maximum number of tests that should be ordered, and a tendency for learned policies to order more tests than necessary. Nonetheless, guided by our outcome function $g(t,x)$, our method consistently achieved the best $\mathcal{L}_b$ across all approaches. Moreover, the similarities between the numbers in Table~\ref{min53-tab} and Table~\ref{min10-tab} suggest that most ICU stays in the MIMIC dataset typically include around 5 test data points every 48 hours.

Furthermore, Table~\ref{min10-tab} highlights our method's superiority in conditions more closely mirroring real-world scenarios compared to the traditional physician policy (logging policy), underscoring our method's effectiveness.

Additionally, we compared our approach to an RL method that uses mortality as the reward. Under more realistic settings, the RL method showed diminished performance in terms of cost, clinical relevancy, and utility to clinicians. Moreover, our approach, which does not rely on mortality as a learning signal, achieved comparable or even superior information gain relative to the RL method focused on patient mortality. This comparison further underscores the importance of designing ICU lab test ordering systems that prioritize clinician needs over patient-facing metrics.

\begin{table*}[t]
\caption{Testset performance of prior work and ours policies (with std) with 1 minimum required tests per 48 hours.}
\label{min10rl-tab}
\begin{center}
\begin{small}
\begin{sc}
\begin{tabular}{lcccr}
\toprule
Policy & $\Delta X \uparrow$ & Cost$\downarrow$ & $\mathcal{L}_b^{test} \downarrow$ & info gain$\uparrow$ \\
\midrule
Physician  & 0.15& 0.33 & 1.12   & 1.07\\
\hline
RL (low cost)& - & 0.32 & 2.1 & 1 \\ 
RL (high cost)& - & 0.78 & 5.3 & 1.24 \\ 
Ours(w/o GPS) & 0.21 $\pm$0.001& 0.38 $\pm$0.002 & 0.91 $\pm$0.003 & (1.22) \\
Ours(w GPS)   & 0.18 $\pm$0.001& 0.28$\pm$0.01 & 1.03$\pm$0.004   & (1.09) \\
\bottomrule
\end{tabular}
\end{sc}
\end{small}
\end{center}

\end{table*}

\subsection{HIRID Results }
The High Time Resolution ICU Dataset (HiRID) \cite{hyland2020early} is a publicly available critical care dataset that originates from a collaboration between the Swiss Federal Institute of Technology (ETH Zurich) and the University Hospital of Bern. HiRID contains a rich collection of high-resolution data from patients admitted to the intensive care unit (ICU), designed to support a wide range of research initiatives in critical care medicine and machine learning.

Spanning over several years, HiRID includes data from thousands of ICU stays, offering detailed information on physiological parameters, laboratory test results, treatment interventions, and more. One of the dataset's distinguishing features is its high temporal resolution, providing minute-by-minute measurements for a subset of variables, which enables the development and validation of predictive models that require fine-grained temporal data.

Originally developed to facilitate the prediction of circulatory failure and other critical events in the ICU, HiRID's comprehensive and detailed nature makes it suitable for a broad array of research questions. This includes studies on disease progression, treatment effect analysis, and the development of decision support tools for clinicians. The dataset's structure allows for the exploration of irregular time-series data in medical contexts, making it an invaluable resource for advancing patient care through machine learning and data-driven approaches.

While the MIMIC dataset provides a substantial repository with approximately 57,000 ICU stays, the High Time Resolution ICU Dataset (HiRID) encompasses around 32,000 stays. Despite this smaller number, HiRID compensates with its detailed and frequent collection of patient data, especially concerning vital signs and treatment interventions. This high granularity makes HiRID an exemplary dataset for irregular time-series analysis, offering a dense array of data points and significantly less missingness, which is instrumental in the development of advanced patient forecasting models.

Upon processing the HiRID dataset, we identified a total of 79 features, including a more diverse assortment of Vasopressors compared to those found within the MIMIC dataset. This diversification in data points allows for a more nuanced analysis of patient responses to various treatments. Notably, the mean squared error for HiRID stands at $0.035 \pm 0.003$, a figure that, while higher than that observed in MIMIC, reflects the reduced incidence of missing data within the dataset. Despite this increased error margin, the loss remains commendably low given the sparsity of the data, suggesting that forecasting models $\phi$ developed using HiRID data possess a higher degree of reliability and applicability than those trained exclusively on MIMIC data.

In applying our machine learning methodology to the HiRID dataset, we followed the same procedural rigor as with our MIMIC dataset application. The results from this endeavor align closely with our baseline comparisons, underscoring the robustness and efficacy of our approach as detailed in Tables ~\ref{tab:hirid1} and ~\ref{tab:hirid2}. This consistency across datasets reinforces the validity of our method and its potential for real-world clinical application.

Notwithstanding the successful application and experimentation with HiRID—a dataset derived from a separate patient cohort—our investigation into the robustness of our methodology in the face of data distribution shifts remains in its infancy. The primary challenge lies in reconciling the differing nomenclatures and feature combinations present in each dataset, such as the variations in Antibiotics listed across MIMIC and HiRID. Nevertheless, the ability to corroborate our findings with an additional, independently collected dataset bolsters confidence in the generalizability and practical applicability of our method.

Looking ahead, these preliminary findings lay the groundwork for more exhaustive future investigations into the interoperability of models trained on one dataset and tested on another, particularly between MIMIC and HiRID. Such research will be crucial in assessing the adaptability and versatility of our machine learning strategies across varied clinical datasets, marking a promising direction for subsequent work.
\begin{table*}[t]
\caption{Testset performance for baseline and our learned policies (with std), on HiRID dataset.}
\label{tab:hirid1}
\begin{center}
\begin{small}
\begin{sc}

\begin{tabular}{lccccr}
\toprule
Policy & $\Delta X \uparrow$ & Cost$\downarrow$ & $\mathcal{L}_b^{test} \downarrow$ & $L_{low}\downarrow$ & $L_{up}\downarrow$ \\
\midrule
$\text{Random}_{0.5}$    & 0.47 $\pm$0.01 & 0.49$\pm$0.1 & 4.4$\pm$0.01    & 2.9$\pm$0.01 & 1.5$\pm$0.03 \\
$\text{Random}_{0.75}$   & 0.58 $\pm$0.01 & 0.74$\pm$0.1 & 3.77$\pm$0.01   & 1.66$\pm$0.01 & 2.11$\pm$0.01 \\
LowerBound               & 0.62 & 0.35 & 0      & 0 & 0 \\
UpperBound               & 1.13 & 0.59 & 0      & 0 & 0 \\
Physician                & 0.96& 0.55 & 0.98   & 0.98 & 0 \\
\hline
Ours(w/o GPS)            & 1.08 $\pm$0.01& 0.57 $\pm$0.02 & 0.62$\pm$0.001   & 0.3 & 0.32 \\
Ours(w GPS)              & 1.01 $\pm$0.01& 0.52$\pm$0.01 & 0.89$\pm$0.001   & 0.5 & 0.39 \\
\bottomrule
\end{tabular}

\end{sc}
\end{small}
\end{center}
\vskip -0.2in
\end{table*}

\begin{table*}[t]
\caption{Testset performance of prior work and ours policies (with std), on HiRID datset.}

\label{tab:hirid2}
\begin{center}
\begin{small}
\begin{sc}

\begin{tabular}{lcccr}
\toprule
Policy & $\Delta X \uparrow$ & Cost$\downarrow$ & $\mathcal{L}_b^{test} \downarrow$ & info gain$\uparrow$ \\
\midrule
Physician & 0.96 & 0.55 & 0.98 & 1.03 \\
\hline
RL (low cost)& - & 0.51 & 1.4 & 1 \\ 
RL (high cost)& - & 0.74 & 3.6 & 1.19 \\ 
Ours(w/o GPS) & 1.08 $\pm$0.01& 0.57 $\pm$0.02 & 0.62$\pm$0.001 & (1.17) \\
Ours(w GPS)    & 1.01 $\pm$0.01& 0.52$\pm$0.01 & 0.89$\pm$0.001& (1.05) \\
\bottomrule
\end{tabular}
\end{sc}
\end{small}
\end{center}

\end{table*}

%\newpage
\section{Other Experiments for Policy Evaluation} \label{apx:exp}

In this section, we delve into additional experimental outcomes, emphasizing the comparative analysis of our formulated policies against standard baselines and reinforcement learning (RL) strategies. Detailed outcomes are encapsulated in Table~\ref{fulltestperf} and Table~\ref{fullrl}, showcasing the efficacy of our policies relative to conventional approaches. 

Our observations reveal that the absence of a generalized propensity score (GPS) does not deter the outcome function's capacity to steer the policy towards achieving a minimized loss, albeit with a tendency to favor policies associated with elevated costs.

For the comparison with RL methodologies, we draw upon the framework of \citet{chang2019dynamic}, who employed the final LSTM hidden layer as the state representation $x$. Aligning our data temporally, we adopt this LSTM layer as our state $x$, assessing our laboratory test orders as actions at each temporal step. This approach facilitates the evaluation of our policy's effectiveness through the cumulative gain in information, gauged by the discrepancy in probabilities as per their off-policy model.

The off-policy evaluation metric, predicated on the regression of state-action pairs against the differential in mortality classifier probabilities, aims to minimize the informational exposure to users at testing phases. This raises an intriguing query: Does the ordering of lab tests directly correlate with patient mortality? Such an assumption may inadvertently suggest a disparity in clinical treatment across patients.

\subsection{Exploring Policy Learning without Predicted Future Insights} \label{apx:x_future}

In pursuit of enhanced explainability and adherence to the logical progression of lab test ordering, we instituted a model for forecasting future patient states, enabling our policy to incorporate anticipated future patient conditions. This not only augments explainability but also highlights a significant reduction in the outcome—specifically, a 20-25\% decline in the utility value of the lab utility function $g(t,x)$ and in the bound loss, particularly when the forecasted future is excluded from policy training.

\subsection{Incorporating Real-world Lab Test Costs}

We adjusted our model to reflect actual lab test costs as documented in literature, with values delineated as $[12, 5, 12.36, 18, 9.1, 10, 18.62, 1.5, 18, 1.5]$ in USD. These numbers are suggested by clinicians with prior studies~\cite{kandalam2020inappropriate, spoyalo2023patient}. The normalization of $\alpha_j$ within our outcome function's cost term leverages this cost array. Policies formulated with this real-world cost paradigm have demonstrated an ability to curtail overall expenses by 5-8\% on average. Given an average test cost of \$10.8, a daily ordering volume of 1000 tests could translate into savings of \$500-900, significantly alleviating hospital financial strains and reducing bio-hazardous waste.
 
\subsection{Ablation Study on Outcome Function Components}

A distinctive aspect of our proposed outcome function $g(t,x)$ is its composition, which encapsulates three key dimensions reflective of optimal lab test ordering practices. Our ablation study on these components reveals their substantial influence on policy formulation.

Eliminating the bound component results in extreme policy behaviors: either an all-inclusive ordering approach to maximize $\Delta X$ or a total abstention to minimize costs. Sole reliance on the $\Delta X$ component propels the policy towards maximal ordering, culminating in a peak $\Delta X$ of 4.52 and a bound loss $L^{test}_b = 2.6$. Conversely, prioritizing cost reduction or assigning significant weight to $\beta_2$ leads to a policy of non-ordering, characterized by a zero $L_{up}$ and a maximum $L_{low} = 6.4$.

Thus, the bound term $\mathcal{L}_b$ emerges as pivotal within the outcome function, guiding the policy towards higher cost strategies yet maintaining a threshold (akin to outcomes observed with an Upper Bound policy). Our exploration into the weighting of these terms suggests that a balanced approach yields favorable policy outcomes, though our analysis was confined to integer weight adjustments. Future investigations might benefit from a comprehensive hyperparameter optimization across the $\beta$ coefficients.

\subsection{Ablation Study on Time-series Model with Different Level of Errors}

We want to investigate how prediction errors in the time-series model influence the learned policy. 
To answer this, we ran two ablation studies to test how the time-series model error would affect the evaluation results (reward function $g$):

\begin{itemize}
    \item For each test sample, we choose to generate the predicted next 24 hours either by random OR using our learned time-series model. Then the policy function is provided input with either [48-hour observation, random next 24-hour] or [48-hour observation, learned model predicted next 24-hour]. Using random predictions resulted in the cost metric for the next 24 hours going up 50\%, $L_b$ went up 43\% with delta X went up only 3\%. The big increase of cost and $L_b$ means that given the random prediction as input, our learned policy is ordering more lab tests that deviate from rule-generated tests or the physician (logging) policy shown in the dataset. However, even though more tests are being ordered the information provided to the clinical team was not significantly increased (as indicated by the small increase in $\Delta X$).
    \item We also compared the actions generated with input of the true next 24-hour and the orders generated using the learned time-series model predicted 24-hour. When using the true next 24-hour as input, our policy generates actions that have a 3\% decrease in cost, 6\% decrease in $L_b$ and 5\% increase in $\Delta X$. The small increase in cost and $L_b$ shows the perfect prediction input allows our learned policy to order a bit more tests to provide a little more information (increase in $\Delta X$) to the clinicians.
\end{itemize}
This shows that our learned time-series model is doing a solid but not perfect job on predicting the patient status since using evaluating our learned policy with perfect prediction is not significantly different from evaluating our learned policy with our learned times-series model prediction. 

In conclusion, if the prediction error of the time-series model is high, then the outcome/reward function will show a worse reward than the time-series model with less prediction error.

In part, our utility of this framework does derive from the quality of the time-series prediction and beyond simplistic settings, it is difficult to perfectly characterize the relationship between the two. 

\subsection{Ablation Study on Simulating `Human Error' in the Privileged Information}

We investigate how errors in rule specification affect the policy learning. In our framework, the rules serve as indicators or bounds that define the minimal set of lab tests to be ordered based on the patient’s status. These rules are not meant to be exhaustive or flawless but provide a conservative baseline to ensure safe practice.

Consider two extremes: if we had perfect rules that match the Bayes optimal predictor for the policy function, we could directly solve the problem using those rules. In contrast, if no rules are applied, the policy's search space ranges from ordering no tests at all to ordering all tests observed under the logging policy; in Algorithm \ref{alg:orderbound}, this scenario corresponds to having all $t^{lower}$ vectors set to zero. Any inclusion of rules results in non-zero $t^{lower}$, which restricts the search space and guarantees that certain tests are ordered to mitigate the risk of missing necessary tests.

The above argument suggests that there is a statistical benefit to having rules. So what happens if the rules contain human errors. 

We first define that we say a rule is more conservative if the condition of the rule is more extreme (less likely to trigger the rule and order corresponding tests). In other words, a more conservative rule will yield more 0’s in $t^{lower}$ than a less conservative rule. 

The human error in the rules therefore yields more conservative rules or less conservative rules.

If human error leads to more conservative rules (i.e. all zeros in $t^{lower}$), this is equivalent to not including any rules in our framework, which means our policy is solving causal bandits without side information.If human error leads to less conservative rules (i.e. more 1’s in $t^{lower}$), it would result in over-ordering as in the logging policy.

On the flipside, there are two possibilities for less conservative rules. 
a] Either a less conservative rule would lead to $t^{lower}$ becoming a vector of 1’s, meaning each patient should get every test order after each 48 hours. But this implies that the physicians in the logging policy (current medical practice) are under-ordering lab tests for patients, which contradicts our assumption that logging policy has over-ordering issues and is suboptimal.

b] Or that less conservative rules would lead to $t^{lower}$ having similar 1’s as  $t^*$, means that the clinician defined rules are ordering similar amounts of lab tests as the logging policy. This also implies that the rules recover a similar ordering policy as the physician policy.

To further support the above argument, we perform another ablation study on the rules. We randomly choose three lab tests from our study and remove all the rules related to these three lab tests, meaning $t^{lower}$ will always be 0 for these three lab tests. We trained the policy with our framework and during evaluation, we found that the total number of these three tests ordered is roughly 9\% less than training with their corresponding rules due to the goal of minimizing cost in the reward. This result shows that if we increase the search space for lab orders, with the guide of our reward function, lab tests that are less triggered by each of these rules would be ordered less. This leads to a policy that takes more risks to potentially miss necessary tests.

With the same three lab tests, instead of removing all corresponding rules, we change the rules to ‘suggest to order if the dataset is ordering’, meaning that $t^{lower}$ is always equal to  $t^{upper}$ for these three lab tests. During evaluation, we found that the total number of these three test orders increased 37\% compared to training with our existing rules and the total number of lab orders for other tests decreased 23\%. This result indicates that if we decrease the search space for lab orders, with the guide of our reward function, lab tests that are triggered by rules similar to the logging policy would be ordered more and cause over-ordering to happen. 

Thus, having rules help the policy generate minimal lab orders and having less conservative rules would revert the policy back to the logging policy which errs on over-ordering.

\subsection{Ablation Study on the Use of Privileged Information (Clinical Rules)} \label{apx:rule_ablation}

We assessed the impact of incorporating privileged information—specifically, clinical rules—into our policy learning framework. These rules serve as guidance to ensure safety and informativeness of the lab test orders selected by the learned policy. We performed an ablation study by comparing our original approach (using Algorithm 1 to compute minimal necessary tests,  $t_{lower}$) against two altered conditions:

1. No Rules: 
Instead of using clinical rules to determine $t_lower$, we set $t_lower$ to zero vectors, effectively removing any privileged guidance.

2. Rules as the Logging (Physician) Policy: 
Instead of using clinical rules, we force $t_lower$ to be the same as the logging policy. This simulates having rules identical to the logging policy, which may over-order tests.

\textbf{Setting 1 (No Rules):}

Behavior of the Learned Policy: Without any privileged information, the policy converged to ordering only low-cost tests (e.g., CK, Creatinine) and avoided relatively expensive tests (e.g., ABG, INR, Troponin). This approach effectively minimizes the cost penalty but disregards safety and informativeness.

Performance Metrics:
\begin{itemize}
    \item $L_{upper}$ increased by about 11\%: The policy’s test orders deviate significantly from the physician’s (logging) policy and the clinical rules.
    \item $L_{lower}$ increased by about 56\%: The learned policy’s behavior diverges greatly from the clinical standard, indicating poor adherence to necessary tests.
    \item Cost metric unchanged: The number of tests ordered does not decrease, but the composition shifts toward cheaper tests.
    \item $\Delta X$ decreased by about 7\%: The selected tests provide less informative clinical insight.
\end{itemize}
These results suggest that without privileged rules, the policy exploits the cost structure and may adopt a clinically suboptimal strategy. Further adjusting the cost weight could lead to trivial (zero-test) solutions, which are even less clinically informative.

\textbf{Setting 2 (Rules = Logging Policy):}

Behavior of the Learned Policy: By setting the rules to match the logging physician policy, the learned policy closely mimics the logging policy’s test selection, leading to negligible differences in $L_{upper}$.

Performance Metrics:
\begin{itemize}
    \item $L_{lower}$ increased by about 15\% compared to the logging policy and by about 34\% compared to using clinical rules. Although the learned policy adheres to the logging pattern, it misses critical tests that clinical rules would have suggested.
    \item Cost decreased by about 6\% compared to the policy trained with clinical rules. This indicates fewer tests being ordered overall.
    \item $\Delta X$ decreased by about 17\%, showing that the selected tests are less informative to the clinical team.
\end{itemize}

These findings show that using the logging policy as ``rules” constrains the policy to the baseline ordering pattern, offering limited incentive to discover more appropriate, informative, and cost-effective test combinations.

The ablation study demonstrates that having privileged clinical rules is critical. Without these rules (Setting 1), the policy finds cost-minimizing but clinically uninformative solutions. Using the logging policy as the rule-set (Setting 2) leads to overfitting on the baseline pattern and fails to discover safer and more informative sets of tests. In contrast, integrating clinical rules (privileged information) ensures that the learned policy can explore a broader range of test combinations while maintaining a clinically safe and informative standard.

\section{Method Explanation for Non-ML Audience}

In our work, we've developed a machine learning method aimed at optimizing blood test ordering for ICU patients, making this process more efficient and informed by data. Our goal is to explain this approach in straightforward terms, particularly for clinicians who could integrate this tool into their daily routines, enhancing patient care without getting bogged down by complex mathematical formulas.

Our method uses existing patient data from Electronic Health Records (EHR) to create two key tools: a forecasting model (represented by $\phi$) and a decision support policy ($\pi$). The forecasting model analyzes the past 48 hours of a patient's data to predict their health status over the next 24 hours, including lab results, treatment responses, and vital signs. The decision support policy then uses this prediction to recommend which blood tests should be ordered for the next day for each patient.

Imagine integrating this tool into an ICU setting. The decision support tool, combined with the forecasting model ($\phi$), guides clinicians on which tests to order next. For example, if the policy ($\pi$) suggests ordering a Complete Blood Count (CBC), a clinician might wonder why. By consulting the model's predictions or the patient's recent data, the clinician can see that the patient recently had a blood transfusion, justifying the need for a CBC test. This approach, recommended by our clinical partners, moves away from abstract data representations and instead bases decisions on actual patient data, making the reasoning behind each test clear.

The significant advantage of deploying our method is its ability to recommend a focused set of blood tests that are both cost-effective and clinically relevant, avoiding unnecessary tests while adhering to clinical guidelines. This not only has the potential to lighten the workload for hospital staff but also ensures patient safety is not compromised.

By directly leveraging real patient data for both present and future predictions, our system offers a clear, rational basis for each recommended test, aligning closely with clinical needs and practices. This could revolutionize how clinicians make decisions about patient care, making the process more efficient and grounded in data-driven insight

\end{document}